\documentclass[10pt,journal,letterpaper]{IEEEtran}
\usepackage{cite}
\usepackage{url}

\usepackage{xcolor}
\usepackage{graphicx}
\usepackage{amsmath}
\usepackage{nicefrac}
\usepackage{amssymb,amsthm}

     \newtheorem{lemma}{Lemma}

    \newtheorem{theorem}{Theorem}
    \newtheorem{definition}{Definition}
    
\newcommand {\red}[1]{#1}

\newcommand{\Painleve}{\mbox{Painlev\'{e} $\;$}}

\DeclareMathOperator{\sign}{sign}
\newcommand{\un}[1]{\ensuremath{\, \mathrm{#1}}}

\newtheorem{defn}{Definition}
\newcommand{\seq}{\mbox{$\! \, = \, \!$}}

\newcommand{\vq}{\mbox{${\mathbf{q}}$}}

\newcommand{\vf}{\mbox{${\mathbf{f}}$}}
\newcommand{\vfi}{\mbox{$\varphi$}}

\newcommand{\vp}{\mbox{${\mathbf{p}}$}}

\newcommand{\vP}{\mbox{${\mathbf{P}}$}}

\newcommand{\vy}{\mbox{${\mathbf{y}}$}}

\newcommand{\Ss}{\mbox{${\mathcal{S}}$}}

\newcommand{\vqd}{\mbox{$\dot{\mathbf{q}}$}}

\newcommand{\xd}{\mbox{$\dot{x}$}}

\newcommand{\zd}{\mbox{$\dot{z}$}}
\newcommand{\zdd}{\mbox{$\ddot{z}$}}

\newcommand{\real}{\mbox{$\mathbb{R}$}}

\newcommand{\Poincare}{Poincar\'{e} }

\newcommand{\beq}[1]{ \begin{equation} \label{eq.#1} }
\newcommand{\eeq}{ \end{equation} }

\begin{document}
%
\title{Experimental Verification of Stability Theory for a Planar Rigid Body with Two Unilateral Frictional Contacts}

\author{Yizhar~Or,~\IEEEmembership{Member,~IEEE,}
        and P\'eter~L.~V\'arkonyi
 
\thanks{
This paper has supplementary downloadable material available at
http://ieeexplore.ieee.org, provided by the authors. This includes a movie clip illustrating the experiments, and raw datasets of experimental results formatted as MatLab figure files. This
material is 10 MB in size.}%
 \thanks{
 Y. Or is with the Faculty of Mechanical Engineering,
Technion - Israel Institute of  Technology, Haifa 3200003, Israel. E-mail: izi@technionac.il.
}

\thanks{
P. L. V\'arkonyi is with the Department of Mechanics Materials and Structures, Budapest University of Technology and Economics, H-1111 Budapest, Hungary}
}


\markboth{Accepted for publication by IEEE Transaction on Robotics}%
{Shell \MakeLowercase{\textit{et al.}}: Bare Demo of IEEEtrancls for IEEE Journals}
%

\maketitle

\begin{abstract}

Stability of equilibrium states in mechanical systems with multiple unilateral frictional contacts is an important practical requirement, with high relevance for robotic applications. In our previous work, we theoretically analyzed finite-time Lyapunov stability for a minimal model of planar rigid body with two frictional point contacts. Assuming inelastic impacts and Coulomb friction, conditions for stability and instability of an equilibrium configuration have been derived. In this work, we present for the first time an experimental demonstration of this stability theory, using a variable-structure rigid ''biped'' with frictional footpads on an inclined plane. By changing the biped's center-of-mass location, we attain different equilibrium states, which respond to small perturbations by divergence or convergence, showing remarkable agreement with the predictions of the stability theory. Using high-speed recording of video movies, good quantitative agreement between experiments and numerical simulations is obtained, and limitations of the rigid-body model and inelastic impact assumptions are also studied. The results prove the utility and practical value of our stability theory.

\end{abstract}


\section{Introduction}
\IEEEPARstart{M}{any} robotic systems are based on establishing contacts between bodies, for performing tasks of object manipulation or locomotion. Several characteristic types of contact-based robotic motion exist. In robotic grasping, contacts are typically used to enforce kinematic constraints that immobilize an object with zero relative motion, so contacts are maintained persistent. On the other hand, dynamic tasks such as object juggling and rapid legged locomotion involve intermittent contacts where impacts induce non-smooth transitions in contact states. An intermediate regime uses {\em quasistatic } manipulation and locomotion tasks with non-prehensile contacts. Such motion often relies on unilateral contacts that are maintained in persistent no-slip state imposed by equilibrating forces that satisfy frictional contact constraints. A common example is quasistatic legged locomotion on rough terrain where gravitational load is resisted by contact forces at the feet's supports.

In the regime of quasistatic motion with unilateral contacts, it is of practical importance to consider {\em stability} of multi-contact equilibrium states under disturbances caused by model uncertainties, joint coordination inaccuracies, irregular contact surfaces, and more. Common approaches consider robustness of the solution for equilibrium contact forces under disturbances such as localized elastic deformations at contacts \cite{howard&kumar,varkonyi2015stability} or margins of potential energy \cite{papadopoulos_icra96}. A main limitation of these approaches is assuming persistent contacts without accounting for dynamics under small initial perturbations about equilibrium, that do not necessarily maintain contact constraints. This is close in spirit to the well-known concept of {\em Lyapunov stability} in dynamical systems theory \cite{leine.stability,lyapunov_book,orlov2020nonsmooth}. Analysis of this type of dynamic stability in multi-contact systems is challenging, since any small initial perturbation of displacements and velocities immediately induce response governed by hybrid dynamics involving non-smooth transitions between contact states and impacts \cite{brogliato,teel_tutorial09}. Such systems often involve complicated phenomena such as solutions with {\em Zeno behavior} \cite{or&ames_tac2011,zhang.zeno.hs} and more rarely, {\em \Painleve paradox} where friction-dominated solution is either indeterminate or inconsistent \cite{champneys2016painleve,or.RCD2014,leine2002}.

A key component in analysis of motion under intermittent contact is {\em impact} - a discontinuous jump in velocities, which is based on assumption of rigid-body contact and lumping the elastic-plastic contact interaction during the abrupt process of collision into an instantaneous change \cite{stronge_book}. Despite the complicated nature of collision mechanics, several rigid-body models of impact exist, with limited levels of accuracy and reliability. In a single-contact frictionless impact, a common assumption is constant coefficient of kinematic restitution of the normal velocity, as in the well-known bouncing ball example and its variations \cite{luck1993bouncing,leine2012global}. This model naturally leads to {\em Zeno solutions}, which contain an infinite sequence of exponentially decaying impacts, accumulating in a finite amount of time \cite{or&ames_tac2011,zhang.zeno.hs}. Single-contact impact under Coulomb's friction is more complicated and is treated using several models, e.g. \cite{chatterjee&ruina_1998,mason&wang_impact} . Rigid-body impacts in multiple-contact systems is an even more complicated issue \cite{brogliato,glocker1995multiple,ivanov1997impact,remy2017ambiguous}, 
{which has been analyzed in extensively in the context of rocking motion \cite{brogliato2012rocking,dimitrakopoulos2012revisiting}, as well as of granular chains \cite{nguyen2019multiple}.}

Incorporating impacts and contact states transitions into stability analysis of multi-contact equilibria, Or and Rimon \cite{or&rimon.icra08b} analyzed the hybrid dynamics of a planar rigid body with two frictional contacts assuming frictionless impacts with kinematic restitution. They derived highly conservative conditions for finite-time Lyapunov stability of frictional equilibrium states, where the perturbed solutions converged to a nearby two-contact equilibrium via Zeno sequences of single-contact or alternating two-contact impacts. The work \cite{leine.NODY2008} studied {attractivity} of {{\em continuous sets of frictional equilibrium points}} of planar mechanical systems under multiple frictional contacts, assuming impact laws under kinematic restitution coefficients. Their stability conditions are based on choosing total mechanical energy as a Lyapunov function guaranteeing that the solution converges to the equilibrium set while staying close to the minimum of potential energy. This is fundamentally different from Lyapunov stability of a specific equilibrium point which is embedded within a continuous set of equilibria, where the solution is required to stay within a bounded neighborhood of that particular point, while a minimum of potential energy may be located further away from it.

A simplifying model which is often realistic for contacting surfaces made of rough compliant material such as rubber pads is {\em inelastic impact}, which assumes that the relative normal velocity vanishes immediately after collision {\cite{brogliato,stronge_book}}. In multi-contact or multi-body systems, this model still leads to rich and complex hybrid dynamics {\cite{chatterjee2002persistent}}. Posa et al \cite{posa.stability.tac2016} have proposed a computational approach for proving Lyapunov stability of equilibria under multiple frictional contacts and inelastic impacts using iterative construction of sum-of-square Lyapunov functions for estimating region of attraction. The main limitations of \cite{posa.stability.tac2016} are conservatism of the stability conditions, and the lack of criteria for identifying {\em instability} of frictional equilibria. The earlier work of Or and Rimon \cite{or&rimon.icra08a} has identified an instability mechanism termed {\em ambiguous equilibrium} where several contact states are simultaneously feasible. The later work of V{\'a}rkonyi et al \cite{Varkonyi.icra2012} has analyzed another instability mechanism called {\em reverse chatter} \cite{nordmark.ima2010} which is a Zeno-like sequence of inelastic impacts and contact-state transition with exponentially-diverging magnitude. In a followup unpublished work by the same authors of \cite{Varkonyi.icra2012}, this phenomenon has been observed experimentally for the first time in a mechanical system, using a simple prototype of a rigid ``biped'' on a slope, after imposing a slight initial perturbation from frictional equilibrium state. Our recent joint work \cite{varkonyi2017lyapunov} presented theoretical analysis of finite-time Lyapunov stability for a planar rigid body with two frictional contacts and frictional inelastic impacts. The analysis in \cite{varkonyi2017lyapunov} reduced the hybrid dynamics of the system in close vicinity of an equilibrium state to a scalar  \Poincare map $R$ and scalar magnitude-growth function $G$, which together encompass the entire response and contact state transitions. Under specific restrictions called {\em persistent equilibrium}, the work \cite{varkonyi2017lyapunov} derived theoretical conditions for stability and instability of frictional two-contact equilibria based on properties of the semi-analytic functions $R$ and $G$, and showed how stability can depend on structural parameters such as friction coefficients and center-of-mass location relative to contact positions.

The goal of this work is to present, for the first time, an experimental demonstration of our stability theory from \cite{varkonyi2017lyapunov}. Motivated by the previous unpublished followup work of \cite{Varkonyi.icra2012}, our experimental setup consists of a rigid ``biped'' with variable structure, which is perturbed from frictional two-contact equilibrium state on an inclined plane. We first present  extension of our theoretical analysis in \cite{varkonyi2017lyapunov} to account for a relaxed notion called {\em weakly-persistent equilibria} and also derive a simpler stability condition. Both modifications cover cases which are relevant to actual properties of our experimental biped and contact geometry. Upon shifting the biped's variable center-of-mass, our theoretical predictions indicate changes between the two instability mechanisms towards stability. These stability transitions are demonstrated experimentally, and high-speed camera recording enables tracking the biped's motion for quantitative comparison with theoretical simulations, as well as assessing the validity of our rigid-body model assumptions. We find excellent qualitative and good quantitative agreement between the theoretical predictions and experimental measurements, and conclude that our model slightly underestimates stability, where discrepancies are mainly due to added energy dissipation caused by damped elastic vibrations and footpads' compression during impacts. The results demonstrates the utility and practical value of our stability theory.

The manuscript is organized as follows. The next section presents the statement and mathematical formulation of the problem. Section \ref{sec:review}  reviews the stability theory from \cite{varkonyi2017lyapunov} while Section \ref{sec:newtheory} presents some extensions of the theory and simulation examples. Section \ref{sec:experiments} presents the experiments and analysis of their results. Finally Section \ref{sec:conclusions} presents a concluding discussion.


\begin{figure*}[!t]
\centering
\includegraphics[width=0.8\textwidth]{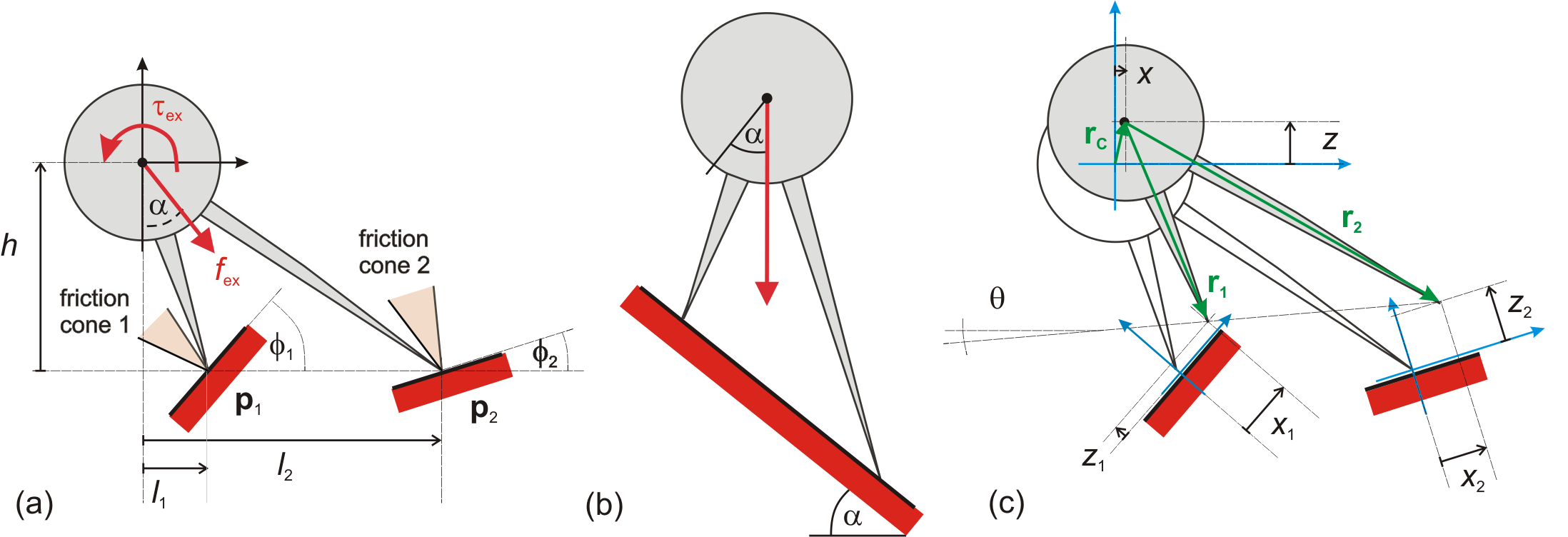}
\caption{(a): Parameters of a rigid body on two frictional point contacts. (b): The special case of body on a slope under gravity (c): {coordinate frames and} state variables.}
\label{fig:notation} 
\end{figure*}

\section{Problem statement, notation} \label{sec:notation}
We analyze the stability of equilibria of a two-dimensional model of a rigid body making contact with a rough, rigid surface at points $\mathbf{p}_1,\mathbf{p}_2$, under constant external loads (Fig. \ref{fig:notation}). The loads as well as the positions and directions of the supports are arbitrary. 
Model parameters (Fig. \ref{fig:notation}a) include the mass $m$ and radius of gyration $\rho$ of the object, the geometric parameters $h$, $l_i$ ($i=1,2$) specifying the locations of the contact points $\mathbf{p}_i$ relative to the center of mass (CoM), $\phi_i$ representing the directions of the contact normals, and $\mu_i$ the Coulomb friction coefficients (static and dynamic coefficients are assumed to be equal). The total external load is expressed as net force vector of magnitude $f_{ex}$ and angle $\alpha$ and net torque $\tau_{ex}$.  
The experimental results presented in this paper focus on objects on slopes under gravity, i.e. $\phi_1=\phi_2=0$, $\tau_{ex}=0$. In this case, $\alpha$ becomes the slope angle (Fig. \ref{fig:notation}b).

Continuous motion of the object is governed by the Newton-Euler equations of rigid body dynamics:
 \begin{align}\label{eqcontinuousmotion}
 \begin{split}
m\ddot{\mathbf{r}}_c&=\mathbf{f}_{ex}+\mathbf{f}_{1}+\mathbf{f}_{2}\\
m\rho^2\ddot{\theta}&=\tau_{ex}+\langle\mathbf{r}_1\times\mathbf{f}_{1}+\mathbf{r}_2\times\mathbf{f}_2\rangle
\end{split}
\end{align}
where $\mathbf{r}_c=(x,z)$ is the position vector of the center of mass, $\theta$ is the rotation angle of the object, $\mathbf{r}_i$ are vectors pointing from the center of mass to the vertex points (Fig. \ref{fig:notation}c), $\mathbf{f}_i$ are contact forces, {and $\langle...\rangle$ denotes the out-of plane component of a vector}.

Let $x_i$, and $z_i$ denote tangential and normal displacements of the vertex points with respect to the initial equilibrium configuration (Fig. \ref{fig:notation}c). Throughout the analysis, we use the set of generalized coordinates $\mathbf{q}=(z_1,z_2,x_2)^T$  and its time derivative $\dot{\mathbf{q}}$ to represent states of the object. The initial equilibrium corresponds to $\vq=\dot{\vq}=0$. The equations of motion \eqref{eqcontinuousmotion} {can be} transformed into the new variables 
{in order to obtain an equation of the form
\begin{align}
\ddot{\mathbf{q}} = \mathbf{F}(\mathbf{q},\dot{\mathbf{q}},\mathbf{f}_1,\mathbf{f}_2).
\label{eq:continuousmotion-brief} 
\end{align}
As we point out below, it is sufficient for local stability analysis of the equilibrium state to use the approximation 

$\ddot{\mathbf{q}} \approx \mathbf{F}(0,0,\mathbf{f}_1,\mathbf{f}_2)$.
The explicit form of this formula is derived in the Appendix.}
 
The contacts are assumed unilateral, and contact forces obey Coulomb law: $|f_i^x| \leq \mu_i f_i^z$ where $f_i^{z}$, and $f_i^{x}$ are normal and tangential components of $\mathbf{f}_i$. This implies that each contact force vector $\mathbf{f}_i$ must lie within a friction cone centered about the contact normal, see Fig. \ref{fig:notation}a. 
The {unilateral frictional contact} relations can be formulated as:
\begin{align}\label{eqcomplcontinuous}
\begin{split}
f_i^{z},z_i & \geq 0 = f_i^{z} \cdot z_i \\
f_i^{z}, \zd_i & \geq 0 = f_i^{z}\cdot\zd_i \mbox{, if } z_i \seq 0\\
f_i^{z} ,\zdd_i &\geq 0 = f_i^{z}\cdot \zdd_i  \mbox{, if } z_i\seq \zd_i \seq 0\\
f_i^{x}&= - \mu_i {\rm sgn}(\xd_i) f_i^{z}
\end{split}
\end{align}
Here, \emph{sgn} denotes the set-valued sign function. We exclude all cases in which 
 non-uniqueness or non-existence of the forward solution in a non-static state (i.e. \Painleve paradox \cite{champneys2016painleve}) occur. {Note that the last equation in \eqref{eqcomplcontinuous} can also be formulated using complementarity relations \cite{brogliato,glocker1993complementarity}.}

Impacts are modeled as instantaneous velocity jumps. The normal ($\hat{f}_i^{z}$) and tangential ($\hat{f}_i^{x}$) contact impulses and post-impact normal ($\zd_i^+$) and tangential ($\xd_i^+$) velocities are assumed to satisfy the relations
\begin{align}\label{eqcompl.impact}
\begin{split}
\hat{f}_i^{z},z_i &\geq 0 = \hat{f}_i^{z}\cdot z_i \\
\hat{f}_i^{z},\zd_i^+ &\geq 0 = \hat{f}_i^{z}\cdot\zd_i^+  \mbox{, if } z_i \seq 0\\
\hat{f}_{i}^x&= - \mu_i {\rm sgn}(\xd^+_i) \hat{f}_i^{z}
\end{split}
\end{align}
for $i=1,2$.
The second equation means that impacts are ideally inelastic, furthermore each point involved in a double impact ($z_1=z_2=0$) is either passive (with  $\dot z_i^+ \geq\hat{f}_i^{z} = 0$) or active and behaves according to the requirement of inelasticity ($\hat f_i^{z} \geq \zd_i^+ = 0$). The last equation is Coulomb's law at the impulse level with the same coefficients as continuous motion. The requirements established above yield a unique solution, except for the rare non-uniqueness for some double impacts\cite{remy2017ambiguous}. For that case, the model is complemented with an a priori order of preference between various types of impacts: impact with two active points preferred over impacts with only one active point; and sticking impact preferred over slipping. {Frictional impact models using Newtonian coefficient of restitution may be energetically inconsistent \cite{stronge_book,brogliato}, however such a problem does not appear here due to the fact that impacts are ideally inelastic.}

We use a slightly modified version of classical Lyapunov stability as proposed in our prior work \cite{varkonyi2017lyapunov}. This condition ensures that the response of the object to any sufficiently small admissible (i.e. penetration-free) state perturbation results in finite-time motion staying in a small neighborhood of the initial equilibrium and ending at another nearby equilibrium state. Formally, we introduce the distance metric
\beq{Delta}
\Delta(\vq,\vqd)=\max\left(\sqrt{z_{1}},\sqrt{z_{2}},\sqrt{|x_{2}|},|\dot{z}_{1}|,|\dot{z}_{2}|,|\dot{x}_{2}|\right).
\eeq
and the definition  
\begin{defn}
Let $\vq \seq 0$ be an equilibrium configuration of a planar rigid body on two frictional contacts. This configuration is called \textbf{finite-time Lyapunov stable (FTLS)} if for every arbitrarily small $\epsilon > 0$ there exists $\delta>0$ such that for any kinematically admissible initial position-and-velocity perturbation $(\vq(0),\vqd(0))$ that satisfies
$\Delta(\vq(0),\vqd(0))<\delta $,
 the emerging motion $\vq(t),\vqd(t)$ satisfies $\Delta(\vq(t),\vqd(t))<\varepsilon $ for all $t>0$. Moreover, the solution must reach a static equilibrium configuration where $\vqd \seq 0$ in a finite time $t_f$ that satisfies $t_f < \varepsilon$.
\end{defn}

\section{Review of previous analysis}  \label{sec:review}

We now review the analysis and the main results of \cite{varkonyi2017lyapunov}.

\subsection{Consistency analysis of contact modes}

We follow the hybrid dynamics approach, which is less compact than the {mathematically equivalent} complementarity formulation \cite{glocker1993complementarity}, yet it offers deeper insight into the \red{detailed course of the motion by enumeration 
of contact modes and transitions between them. All these contact modes are encapsulated in the complementarity formulation in a unified way.} During continuous motion, each contact point is in one of 4 modes including free motion (F), sticking (S) or slipping in the positive or negative $x_i$ directions (P,N). These modes corresponds to different values of ${\rm sgn}(\dot x_i)$ as well as different combinations of equalities and inequalities in \eqref{eqcomplcontinuous}. Accordingly, one needs to distinguish between $4^2=16$ possible two-contact modes for the body, each of which is characterized by a two-letter word from the alphabet $\lbrace F,S,P,N\rbrace$. If we exclude the degenerate geometric arrangements corresponding to $\cos\phi_i=0$, then exactly {ten} of these are kinematically admissible due to the constraints of rigid body kinematics. For example, exactly one of the $PP$, and $PN$ modes is kinematically admissible. Mode $PS$ is never admissible, whereas mode SS corresponds to static equilibrium and it is always kinematically admissible.
%
%

Each contact mode is associated with equality and inequality constraints \cite{varkonyi2017lyapunov}. For each mode $\mathcal{M}$, there are altogether 4 equality constraints in the variables $f_i^{z},f_i^{x},\ddot z_i,\ddot x_i$. These can be combined with \eqref{eq:continuousmotion-brief} 
in order to determine the generalized acceleration of the system $\ddot\vq^\mathcal{M}(\vq,\vqd)$ and the corresponding contact forces $\vf_i^\mathcal{M}(\vq,\vqd)$ as functions of the state variables. As the model is used to investigate motion in a small neighborhood of the initial state $\vq=\vqd=0$, the work \cite{varkonyi2017lyapunov} introduced the \emph{zero-order dynamics} (ZOD) approximation. Under the ZOD, contact force and acceleration values are approximated in each contact mode by constant values  $\ddot \vq^\mathcal{M} := \ddot\vq^\mathcal{M}(0,0)$, and $\vf_i^\mathcal{M} := \vf_i^\mathcal{M}(0,0)$ {evaluated at the initial equilibrium state $\mathbf{q}=\dot{\mathbf{q}}=0$:}. The velocity jumps associated with impacts are calculated by using similar approximations. All results summarized below refer to motion of the object under ZOD. 

{The consistency analysis of contact modes in general multi-contact systems is highly demanding. For the uniqueness of a consistent mode, only conservative conditions are available \cite{blumentals2016contact,brogliato2020contact}. Our case with only two contacts is simpler and a brute-force check is tractable.
The inequality constraints are tested for each contact mode one by one in order to decide } which contact mode is consistent in a given state. We exclude systems with marginally consistent contact modes where consistency depends on higher-order terms neglected by the ZOD. 
Once these non-generic situations are eliminated, consistency analysis under ZOD becomes straightforward, and it uncovers some important properties of the system:
\begin{itemize}
\item If SS is consistent in the initial state ($\vq=\vqd=0$), then that state is a {\bf frictional equilibrium}
\item If any other mode is also consistent in the initial state, then that state is an {\bf ambiguous equilibrium}. As we point out below, ambiguity always implies instability. 
\item If every non-static ($\vqd\neq 0$) state in a small neighborhood of an equilibrium has a unique consistent contact mode  (and thus unique forward acceleration) under the ZOD, then the equilibrium is said to be free of Painlev\'e's paradox, or simply {\bf non-Painlev\'e}. 
As it has been pointed out, our analysis is restricted to non-Painlev\'e equilibria.
\item Consider the set of non-static states with two sustained contacts ($z_i=\dot z_i=0$, $\dot x_i\neq 0$). If for all of these states, there exists a unique consistent mode, which belongs to the set $\lbrace PP, NN, PN,NP\rbrace$, then the equilibrium configuration $\vq \seq 0$ is called {\bf persistent equilibrium}. Persistence  outrules transitions from states with two sustained contacts into states with less than two contacts, which facilitates the analysis. In \cite{varkonyi2017lyapunov} the stability analysis was restricted to persistent equilibria. In this paper the requirement of persistence is replaced by a less restrictive one, which allows the theoretical investigation of our physical experiments.
\end{itemize} 

\begin{figure}[!t]
\centering
\includegraphics[width=0.5\columnwidth]{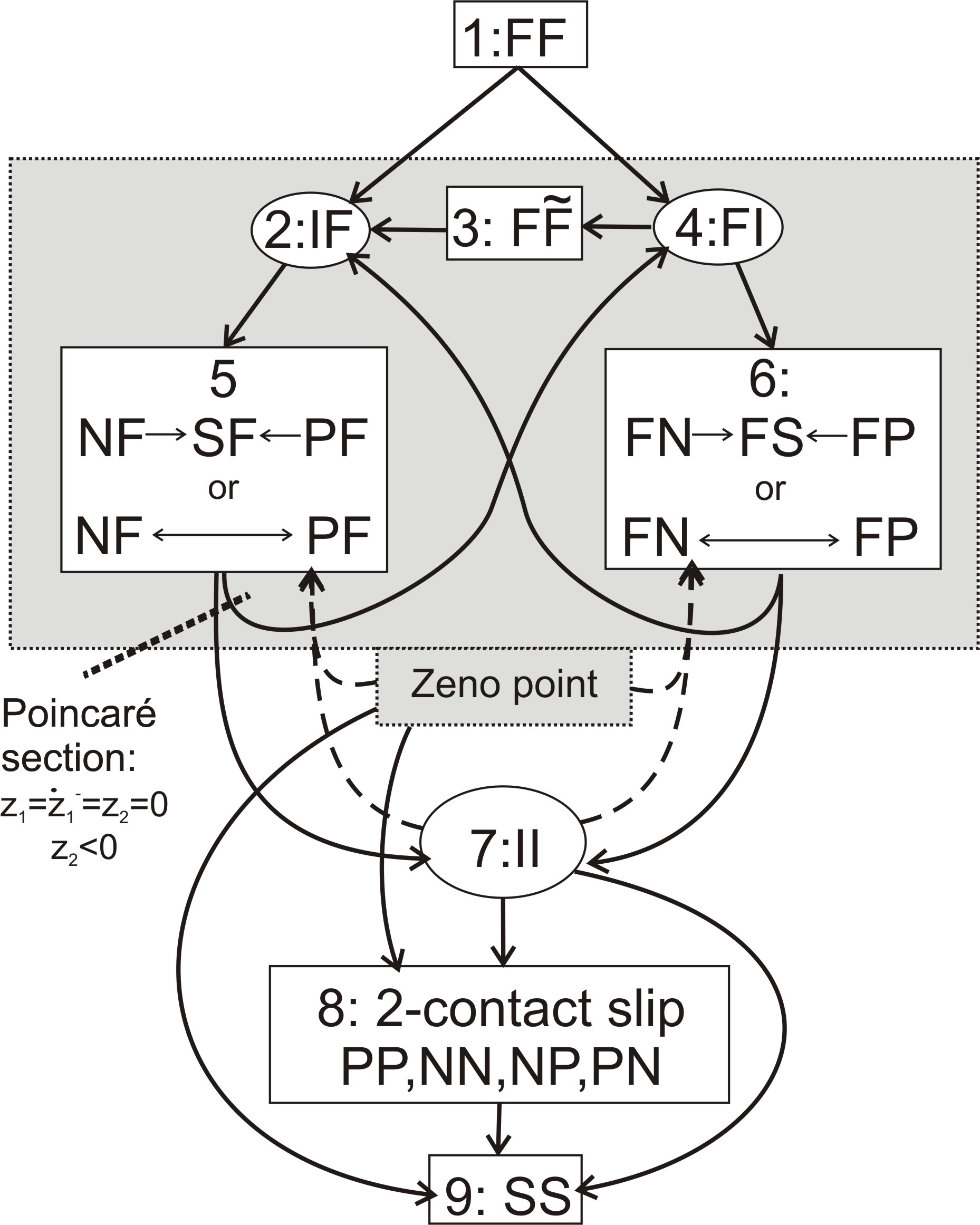}
\caption{Transition graph of the hybrid dynamics under ZOD in the neighborhood of an unambiguous equilibrium. Rounded nodes denote instantaneous impacts, and rectangular nodes represent continuous-time motion.}
 \label{fig:graph}
\end{figure}

\subsection{Hybrid dynamics}\label{sec:hybriddyn}

In \cite{varkonyi2017lyapunov}, we performed a detailed analysis of the hybrid dynamics induced by a small perturbation of an unambiguous equilibrium. 
The results of that analysis are illustrated by the contact mode transition graph in Fig. \ref{fig:graph}. Rectangular nodes with white background represent continuous-time motion in various contact modes, and rounded nodes with white background denote instantaneous impacts. Node 1 corresponds to contact-free motion with arbitrary initial condition, which may follow immediately after the initial perturbation. In contrast, the label $F\tilde{F}$ at node 3 indicates contact-free motion with a special initial condition $z_2=\dot z_2=0$, which may occur multiple times during the motion. {The asymmetry of the graph is caused by how the labels $\mathbf{p}_1$, $\mathbf{p}_2$ are assigned to contact points } (see details in \cite{varkonyi2017lyapunov}). 
Nodes IF and FI mean single-point impacts while the other contact point is separated. Node II means a double impact with both points in contact.

The dashed arrows of the graph correspond to transition which are only possible in the neighborhood of a non-persistent equilibrium, and thus they are eliminated during the analysis of persistent equilibria. The main challenge of the analysis in \cite{varkonyi2017lyapunov} is dealing with closed loops of the transition graph within the area highlighted by grey background, which corresponds to infinite sequences of impacts and continuous motion. Such a sequence may cause unbounded divergence from the initial  state (implying instability of the equilibrium), or it may correspond to gradually decaying motion of infinitely many cycles completed in finite time. The second option corresponds to so-called \emph{Zeno behavior}, which is a generic phenomenon in hybrid mechanical systems. In the present system under analysis, Zeno behavior implies converges to a state with two sustained contacts through an infinite sequence of impacts. This is represented in the figure by a special node labelled as "Zeno point". If the system escapes the infinite loop for all possible initial
perturbations, either through a Zeno point, or through a double impact (node 7), then it eventually reaches static equilibrium (node 9), hence it is FTLS. Accordingly, the stability analysis boils down to analyzing the motion while the system undergoes cyclic transition cycles within the loop denoted by gray region in the graph.

We also found in \cite{varkonyi2017lyapunov} that the system can undergo at most two consecutive impacts before it crosses  
a three-dimensional \Poincare section (see Fig. \ref{fig:graph}) 
$\Ss$ corresponding to pre-impact states, in which vertex point 2 reaches the contact surface, while point 1 stay has sustained contact as:
\beq{Psec}
\Ss = \{(\vq,\vqd) \in \real^{6}: z_1=z_2=0,\; \zd_1^-=0 \mbox{ and } \zd_2^-<0 \}. \eeq
$\Ss$ is parametrized by pre-impact values of three variables, augmented in the vector $\vy = (x_2,\zd_2^-,\xd_2^-)$. The associated  \Poincare map is defined as $\vP: \Ss \to \Ss$ which maps a point $\vy$ of initial conditions on the section $\Ss$ to the values at the next time that the state of the solution trajectory crosses $\Ss$. 
That is, the map induces a discrete-time dynamical system $\vy^{(k+1)} = \vP(\vy^{(k)})$ of the pre-impact states once every step of this impact event. 
We also found that
$\vP$ 
may be undefined for some initial conditions $\vy \in \Ss$, where the solution 
reaches static equilibrium (node $SS$)
without intersecting $\Ss$ again. 

Invariance properties of the ZOD enable (see \cite{varkonyi2017lyapunov}) to define a reduced scalar \Poincare map
\beq{R} 
\varphi^{(k+1)} = R(\varphi^{(k)}). 
\eeq
 in the {angle of impact} 
\beq{varphi} \varphi = \tan^{-1}\left(\frac{\xd_2^-}{|\zd_2^-|}\right). 
\eeq
over the open interval $I:(\pi/2,\pi/2)$.
{Note that $\varphi\rightarrow\pm\pi/2$ corresponds to a \emph{grazing impact}, i.e. $\dot{z}_2^-\rightarrow 0$.}
A related scalar function called {\em growth map} $G: I \to \real_+$,
{is} defined as 
\beq{G} G(\varphi^{(k)}) = \frac{|\zd_2^{(k+1)}|}{|\zd_2^{(k)}|}. 
\eeq
Similarly to $\mathbf{P}$, the reduced maps $R$, $G$ may also be undefined at some parts of $I$. The stability and instability conditions of \cite{varkonyi2017lyapunov} are expressed in terms of the maps $R$ and $G$. 

As we will see, fixed points of $R$, i.e. those values $\varphi^*\in I$ with $\varphi^*=R(\vfi^*)$ as well as the corresponding growth rates $G^*:=G(\varphi^*)$ are of special importance. Such points correspond to cyclic motion of the system, during which it undergoes scaled versions of the same trajectory repeatedly. The duration of subsequent cycles as well as the peak velocities scale by a factor of $G^*$, and subsequent peak values of $z_i$ scale by a factor of $(G^*)^2$.


\subsection{Sufficient conditions of instability}

In some cases, proving the instability of an equilibrium is straightforward. In particular, it was proven  by \cite{or2008hybrid} that
\begin{theorem}[\cite{or2008hybrid}] \label{thm.strong} Ambiguous equilibrium states {\bf do not} possess finite-time Lyapunov stability.
\end{theorem}
This result follows from the fact that each ambiguous equilibrium can be perturbed in such a way that it moves in its consistent non-static contact mode and diverges from the equilibrium state monotonically (under he ZOD) without impacts or contact mode transitions. In the current paper, we will present experimental evidence of the instability of ambiguous equilibria.

The stability analysis of unambiguous equilibria is more subtle. In principle, unambiguity implies that whenever one or two contacts lift off in response to a small perturbation, the system must undergo hybrid motion involving episodes of continuous motion in various contact modes interleaved by impacts. In a model consisting of ideally rigid components, this motion may converge through an infinite sequence of impacts to another nearby equilibrium in finite time (Zeno behaviour) or the system may diverge through a sequence of impacts exponentially from its initial state (reverse chatter \cite{nordmark2011friction}). These complex forms of motion were succesfully analyzed by \cite{varkonyi2017lyapunov} with the aid of the \Poincare maps introduced in Sec. \ref{sec:hybriddyn}.

In order to prove instability of an unambiguous equilibrium, it is sufficient to find one particular type of infinitesimal perturbation, for which the system diverges from equilibrium. This simple observation leads immediately to a simple sufficient condition of instability:
\begin{theorem}[\cite{varkonyi2017lyapunov}] \label{thm.instability_simple}
If the reduced \Poincare map $R(\vfi)$ associated with the ZOD in the neighborhood of an equilibrium configuration has a fixed point $\varphi^*$, for which the corresponding growth rate is $G^*>1$, then the equilibrium configuration is not FTLS.
\end{theorem}
Even though the theorem only gives a sufficient condition of instability, we found by examining many examples numerically, that almost all occurrences of instability are captured by this condition. One example of instability captured by Theorem \ref{thm.instability_simple} is presented in Sec. \ref{sec:examples} (example B).

\subsection{Sufficient conditions of stability}


Stability means that the system remains in a small neighborhood of its initial configuration in response to \emph{any type} of infinitesimal perturbation. The stability of persistent equilibria is guaranteed if the growth function is everywhere below 1 as stated by

\begin{theorem} [ \cite{varkonyi2017lyapunov} ] \label{thm.stability_simple}
 If for a persistent equilibrium, $G(\varphi)<1$ for all $\vfi \in I$ where $R(\vfi)$ is defined, then the equilibrium configuration is FTLS.
 \end{theorem}
The proof of Theorem \ref{thm.stability_simple} in \cite{varkonyi2017lyapunov} is based on the observation that $G(\varphi)<1$ implies that impact sequences triggered by perturbations decay exponentially, until the system returns to two sustained contacts. 
Once the system has two sustained contacts, persistence guarantees that both contacts are preserved, and unambiguity implies that slipping on two contacts deccelerates until the systems stops at a nearby equilibrium configuration.

In many cases, an equilibrium is stable but $G(\varphi)<1$ holds only for some but not for all values of $\varphi$. Then, one can examine the return map $R(\varphi)$ in order to see, which ranges of $\varphi$ are visited repeatedly during the motion of the system. After that, the requirement of $G(\varphi)<1$ can be narrowed down to relevant ranges of $\vfi$. The paper \cite{varkonyi2017lyapunov} introduced the concept of \emph{stable partitions}, which was defined as a partitioning of the  interval $I:(-\pi/2,\pi/2)$ into two subsets: a `safe' set where $G(\vfi<1)$ and a `transient' set which is left by the map $R$ after finite number of iterations. A formal definition of stable partitions is given in the Appendix. The concept of stable partitions allowed us in \cite{varkonyi2017lyapunov} to prove
\begin{theorem}[\cite{varkonyi2017lyapunov}] \label{thm.stability_general}
If for a persistent equilibrium, a stable partition exists, then that equilibrium is FTLS.  \end{theorem}
It should also be noted that the existence of a stable partition appears to be equivalent to requiring $G(\vfi)<1$ over the \emph{non-wandering set} of the map $R$ \cite[Sec. 3.2]{galias2001interval} where `non-wandering set' is a concept widely used in the mathematical theory of maps. The formal proof of this equivalence is beyond the scope of our work.

\begin{figure}[h]
\centering
\includegraphics[width=\columnwidth]{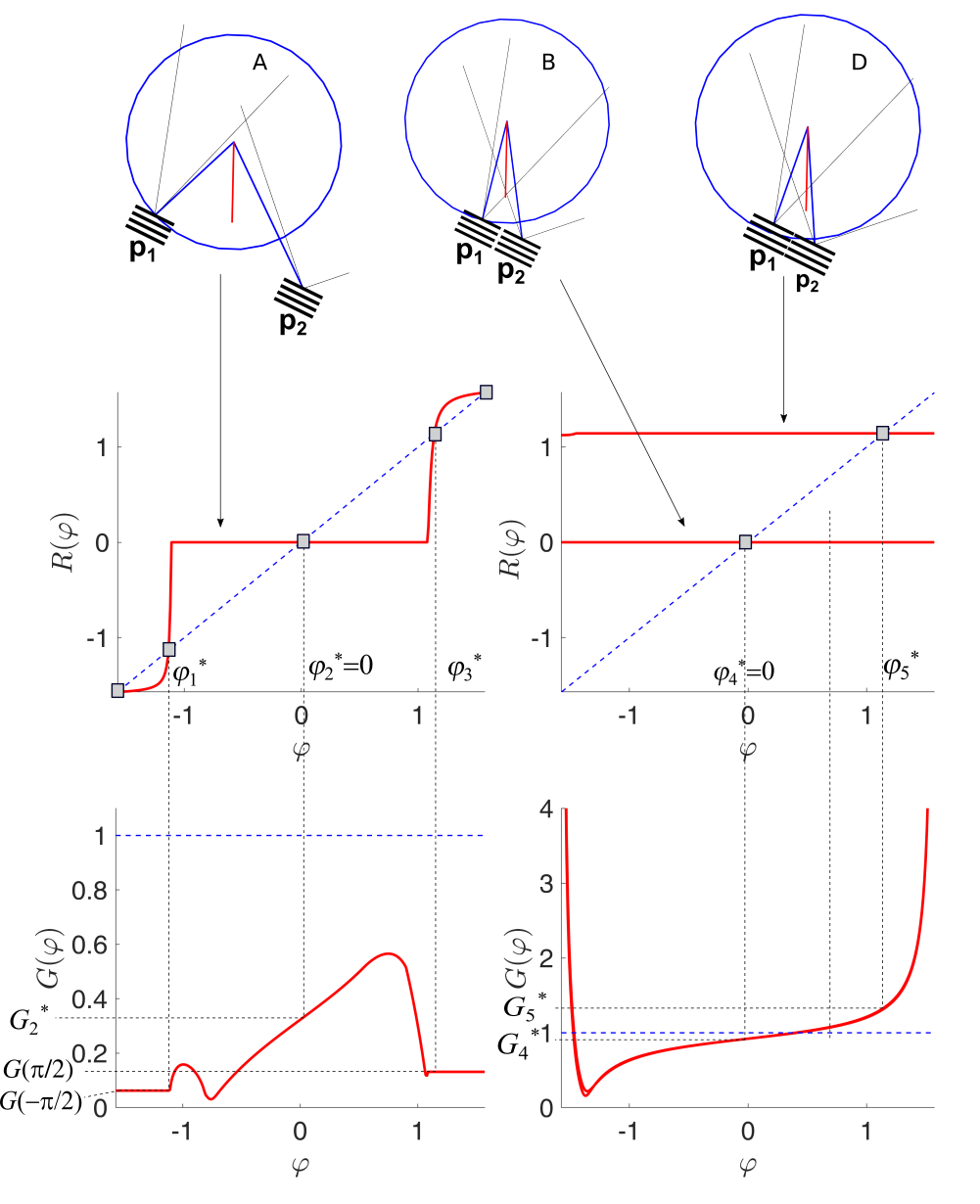}
\caption{Top: {schematic illustrations of t}hree 
biped configurations. Circles denote the biped's radius of gyration $\rho$.  Model parameters are $\alpha=25^\circ,\phi_1=\phi_2=0,\mu_1=0.3150,\mu_2=1,\rho=143.0\un{mm},l_1=-51.2\un{mm},l_2=168.8\un{mm},h=134.1\un{mm}$ for (A), identical values except $\rho=146.9\un{mm},l_1=16.1\un{mm},l_2=76.1\un{mm}$ for (B), and  $\rho=137.9\un{mm}$, $l_1=28.8\un{mm}$, $l_2=88.8\un{mm}$ for (D). Bottom: $R$ and $G$ maps associated with these configurations. Invariant points $\vfi_k^*$ of the $R$ map are highlighted by square markers. Note that the $G(\vfi)$ functions corresponding to B and D are nearly identical, and the $R(\vfi)$ functions are almost everywhere constants (except in a small neighborhood of $-\pi/2$).}
\label{fig:RGmaps}  
\end{figure}

\subsection{Examples}\label{sec:examples}

The results of the theoretical analysis are now presented for three  configurations (Fig \ref{fig:RGmaps}, top). 
All of them correspond to unambiguous equilibria. The $R(\vfi)$ and $G(\vfi)$ maps have been computed numerically for each one and are also depicted in Fig. \ref{fig:RGmaps}. 

\begin{figure*}[t]
\centering
\includegraphics[width=\textwidth]{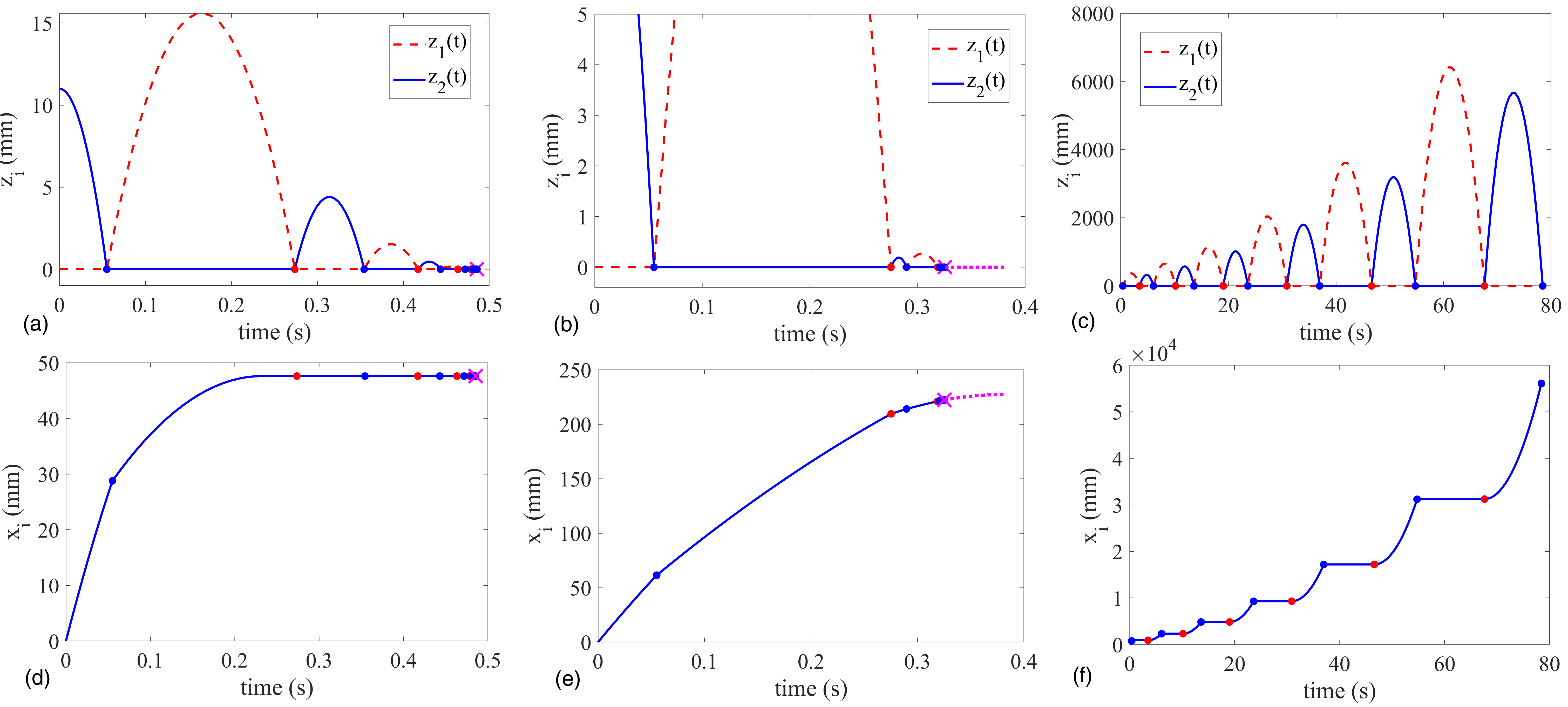}
\caption{Numerical simulation of motion under the ZOD induced by a small initial perturbation. For the persistent, and stable posture A (see Fig. \ref{fig:RGmaps}), the motion sequence converges to an equilibrium for any small initial perturbation. Two possible trajectories are shown in panels (a,d), and (b,e). For each of them, an infinite sequence of impacts terminates at a Zeno point (x marker). In the first case, the Zeno point corresponds to a state of equilibrium, whereas for the second one, it corresponds to finite slip velocity, and additional slip motion (dotted curves) terminates at equilibrium. For the weakly persistent, and unstable equilibrium posture D (see Fig. \ref{fig:RGmaps}), a small initial pertubation results in divergent motion during which the object repeatedly switches between slip motion on point $\mathbf{p}_1$, and roll motion on point $\mathbf{p}_2$ via a sequence of impacts represented by filled circles (panel c,f). }
\label{fig:timeplots} 
\end{figure*}

Configuration A corresponds to an unused configuration of the experimental setup introduced below. Its equilibrium is persistent, and existing stability conditions ensure that it is finite-time Lyapunov stable. In particular, Theorem \ref{thm.stability_simple} is applicable, as we have $G(\vfi)<1$ for all $\vfi\in I$. The corresponding $R$ map 
has three fixed points denoted by $\vfi_1^*,\vfi_2^*=0,\vfi_3^*$, in addition to two asymptotic fixed points at the endpoints $\pm\pi/2$. Among these, $\vfi_1^*$, and $\vfi_3^*$ are repulsive, whereas $\vfi_2^*$, and the endpoints $\vfi=\pm\pi/2$ are attractive, which means that the typical response of the object to a small perturbation under the ZOD may follow several different patterns (after initial transients have died out). Fig. \ref{fig:timeplots}a,b,d,e show displacement-time plots to illustrate two of these alternative forms of motion.Impacts are denoted by filled circles markers in the time plots. Both trajectories converge to static equilibrium in finite time through a Zeno point, denoted by 'x' in the time plots. The first motion sequence corresponds to $\vfi\rightarrow \vfi_2^*$, in which case a decaying sequence of impacts leads to static equilibrium (SS) directly. The second sequence corresponds to $\vfi\rightarrow\pi/2$, in which case the decaying sequence of impacts  is followed by additional two-contact slippage (PP mode) until the object stops (SS mode). Also note the different decay rates corresponding to the values $G_2^*$, and $G(\pi/2)$.  

Configurations B,D of Fig \ref{fig:RGmaps} 
have been tested in our physical experiments. The corresponding theoretical $G$ maps are nearly identical, while the $R$ maps are constant everywhere except for a very small neighborhood of $-\pi/2$. Accordingly, the $R$ maps have a single, globally attractive fixed point $\varphi^*$, which is almost always reached in a single iteration. This property implies that the response of the object under the ZOD is not sensitive to the exact type of the initial perturbation, and initial transients disappear rapidly. Such a property makes these configurations particularly suitable for physical experiments.

Importantly, the equilibria of both configurations are non-persistent, and thus they fall outside the scope of the stability theory previously published in \cite{varkonyi2017lyapunov}. Nevertheless, numerical simulations suggest that configuration B is stable as its response to perturbations  convergence to a nearby two-contact equilibrium through a Zeno sequence of impacts \cite{or2008hybrid}. Cases like this one are our main motivation to develop the generalized stability theory in Sec. \ref{sec:newtheory}. Configuration D is provably unstable by Theorem \ref{thm.instability_simple}, as the $R$ map has a fixed point $\vfi_5^*$ with growth rate $G_5^*>1$. Instability corresponds to divergent motion in response to an initial perturbation. Moreover, the global attractivity of $\vfi_5^*$ under $R$ implies that \emph{any type of perturbation} induces divergence through reverse chatter \cite{varkonyi2012lyapunov}. This type of motion is demonstrated in the time plot of Fig. \ref{fig:timeplots}c,f. 

{In response to small perturbations, configuration B and D undergo similar motion sequences, however delicate differences in the balance of kinetic energy absorption (by impacts and friction) and kinetic energy gain (by downhill motion) make B stable, but D unstable. Notably,} the only difference between configurations B and D is the position of the CoM relative to the vertex points. This makes these configuration suitable for studying transitions from stability to instability experimentally by varying the position of the CoM.

\section{Extension of the stability theory}
\label{sec:newtheory}

Our previous work  focused on persistent equilibria.
However, the experiments presented in the sequel reveal that many configurations of practical interest are indeed non-persistent, similarly to configurations B,D in Fig. \ref{fig:RGmaps}. To address this issue, we now present an extension of the theory to a broader class of equilibria.

Non-persistence has two important consequences. First of all, non-persistent systems may undergo transitions from two contacts ($z_1=z_2=\dot z_1=\dot z_2=0$) to a state with less than two contacts. 
Our systematic investigations reveal that this phenomenon may not occur but for a very small fraction on non-persistent equilibria. Even though our stability theory could be extended to include such systems, we leave this for future work.

Non-persistence has another possible consequence, which occurs much more frequently. All proofs in \cite{varkonyi2017lyapunov} use certain properties of the maps $R(\varphi)$ and $G(\varphi)$, which are true for all persistent systems. The maps associated with non-persistent equilibria may also behave differently. We will address this problem in the sequel. 

In what follows, we generalize Theorem \ref{thm.stability_general} to those non-persistent equilibria, for which transition from two-contact states back to single contact can be excluded despite non-persistence. To this end, we introduce the concept of \emph{weak persistence}:

\begin{figure*}
\centering
\includegraphics[width=0.97\textwidth]{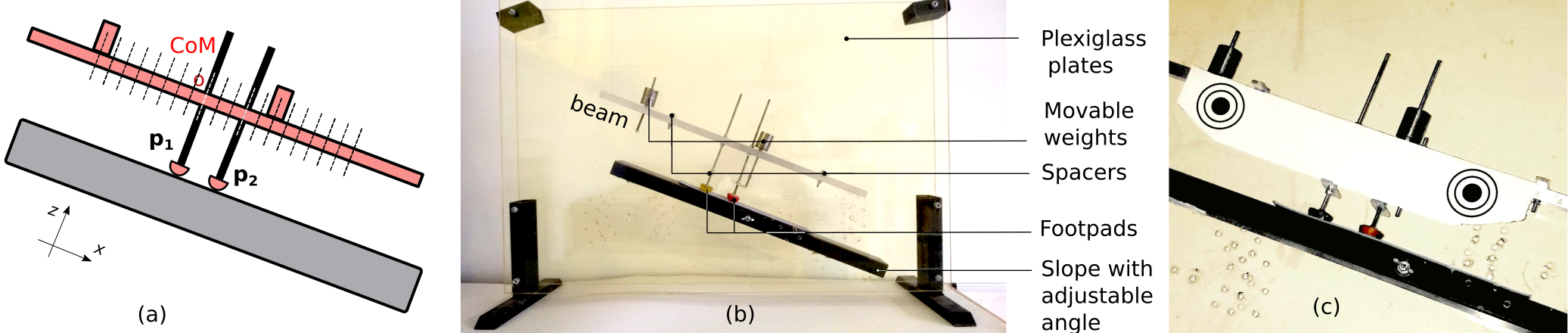}
\caption{{(a,b): Kinematic scheme and photo of the experimental setup. (c):} the biped with attached circular markers for motion tracking.} \label{fig_setup}
\end{figure*}

\begin{defn} \label{def:weakpersistence}
An equilibrium state is \textbf{\emph{weakly persistent}} if it is unambiguous and additionally for both directions of 2-contact slip motion ($z_i=\dot z_i=0$, $\pm\dot x_2 > 0$), at least one of the following conditions is satisfied:
\begin{enumerate}
\item In each state with this slip direction, slipping without separation (PP, NN, PN, or NP mode) is the only consistent contact mode under the ZOD.
\item The corresponding 2-contact slip mode (PP or NP when $+\dot x_2 > 0$, and NN, PN, when $-\dot x_2 > 0$) is not attractive in the following sense:
\begin{enumerate}
\item Transition to that particular mode through a Zeno point  is not possible, i.e. either
\beq{R.pi2:weakpersistence}
|\lim_{\vfi \to \pm \tfrac{\pi}{2}}R(\vfi)|<\pi/2
\eeq
or
\beq{dR.pi2:weakpersistence}
\lim_{\vfi \to \pm \tfrac{\pi}{2}}R'(\vfi)>1
\eeq
\item Nor is it possible to transition to that particular mode through a double slipping impact, i.e. there is no $\varphi$ for which the trajectory includes a slipping double impact with the appropriate slip direction. 
\end{enumerate}
\end{enumerate}
\end{defn}
Note that weakly persistent equilibria include all those  equilibria, where for each of the two possible slip directions, either transition from two-point slip to a state with less than two contacts is impossible (condition 1) or a state of double slip is never reached (condition 2). This set includes all persistent equilibria, which satisfy condition 1 for both slip directions.

With the new definition at hand, we can state a straightforwards extension of Theorem \ref{thm.stability_general}, which is the main theoretical constribution of this paper:

\begingroup
\renewcommand\thetheorem{\ref{thm.stability_general}a}
\begin{theorem} \label{thm.stability_general2}
If for a weakly persistent equilibrium, a stable partition exists, then that equilibrium is FTLS.
\end{theorem}
\endgroup
Indeed the proof of the extended theorem is largely the same as the proof of Theorem \ref{thm.stability_general} in \cite{varkonyi2017lyapunov} with some differences caused by relaxing the requirement of persistence.

First, the original proof in \cite{varkonyi2017lyapunov} used the requirement of \emph{persistent equilibrium} to ensure that once two-contact state is reached, both contacts are maintained until the object stops. The extended theorem requires \emph{weak persistence}, but this property has the same consequence, as we pointed out above.

A second difference in the proof arises from the possibility of special behavior of the maps $R$, and $G$ at the limit points $|\vfi|\rightarrow \pi/2$, which is not possible in the case of persistent equilibria. In order to tackle this issue, we review in the appendix those definitions and proofs used by Theorem \ref{thm.stability_general}, which rely on persistence. Then, we also present all additions necessary to prove the more general Theorem \ref{thm.stability_general2}.

The last theoretical contribution is a corollary of the main stability theorem, which enables a simple and intuitive stability test in many cases:
\begingroup
\renewcommand\thetheorem{5}
\begin{theorem} \label{thm.final}
If for a weakly persistent equilibrium, the associated $R(\vfi)$ function is non-decreasing, and $G(\vfi^*)<1$ for each fixed point $\vfi^*$ of $R$, then that equilibrium is FTLS.
\end{theorem}
\endgroup

The proof of this theorem appears in the Appendix. For non-decreasing $R$ maps, the new theorem is a perfect replacement for Theorem \ref{thm.stability_general2}. It is also notable how Theorem \ref{thm.instability_simple} and Theorem \ref{thm.final} provide a nearly complete stability classification of frictional equilibria.

Numerical computations suggest that objects on a slope always induce monotonic $R$ maps as exemplified by the three configurations of Fig. \ref{fig:RGmaps}. Accordingly, Theorem \ref{thm.final} ensures the stability of configurations A,B.

%

\section{Experiments}  \label{sec:experiments}

We now describe the experiments conducted in order to corroborate our stability theory. The experimental setup (Fig. \ref{fig_setup}) consists of a variable-structure biped with two frictional footpads $\vp_1,\vp_2$ supported in equilibrium on an inclined bar with slope angle {$\alpha=25^\circ$}. The biped consists of an elongated rectangular beam (mass 124~g) with a series of equally spaced drilled holes {(dashed lines in Fig. \ref{fig_setup}.a)}. Two slender legs (40~g) with adjustable lengths and two heavy cylindrical weights (195~g each) are mounted onto the beam's holes in different locations which can be varied in order to control the legs' spacing and center-of-mass (CoM) position. Two hemispheres made of rubber balls are glued at the end of the legs. The biped moves between two vertical plexiglass plates and four spacers ensure that its motion is constrained to the vertical plane. One of the rubber footpads (the ``upper'' one $\vp_1$) is covered with tape in order to reduce its friction. The static friction coefficients of the two feet have been measured in preliminary static tests on the variable slope, and their mean values were found to be $\mu_1=0.315$ and $\mu_2 = 1$. 
In each dynamic experiment, the biped is first placed in two-contact equilibrium on a fixed slope and then slightly perturbed and released in order to test the dynamic response in vicinity of equilibrium. A movie of some experiments is included in the supplementary multimedia extension.

\begin{figure}
\centering
\includegraphics[width=0.9\columnwidth]{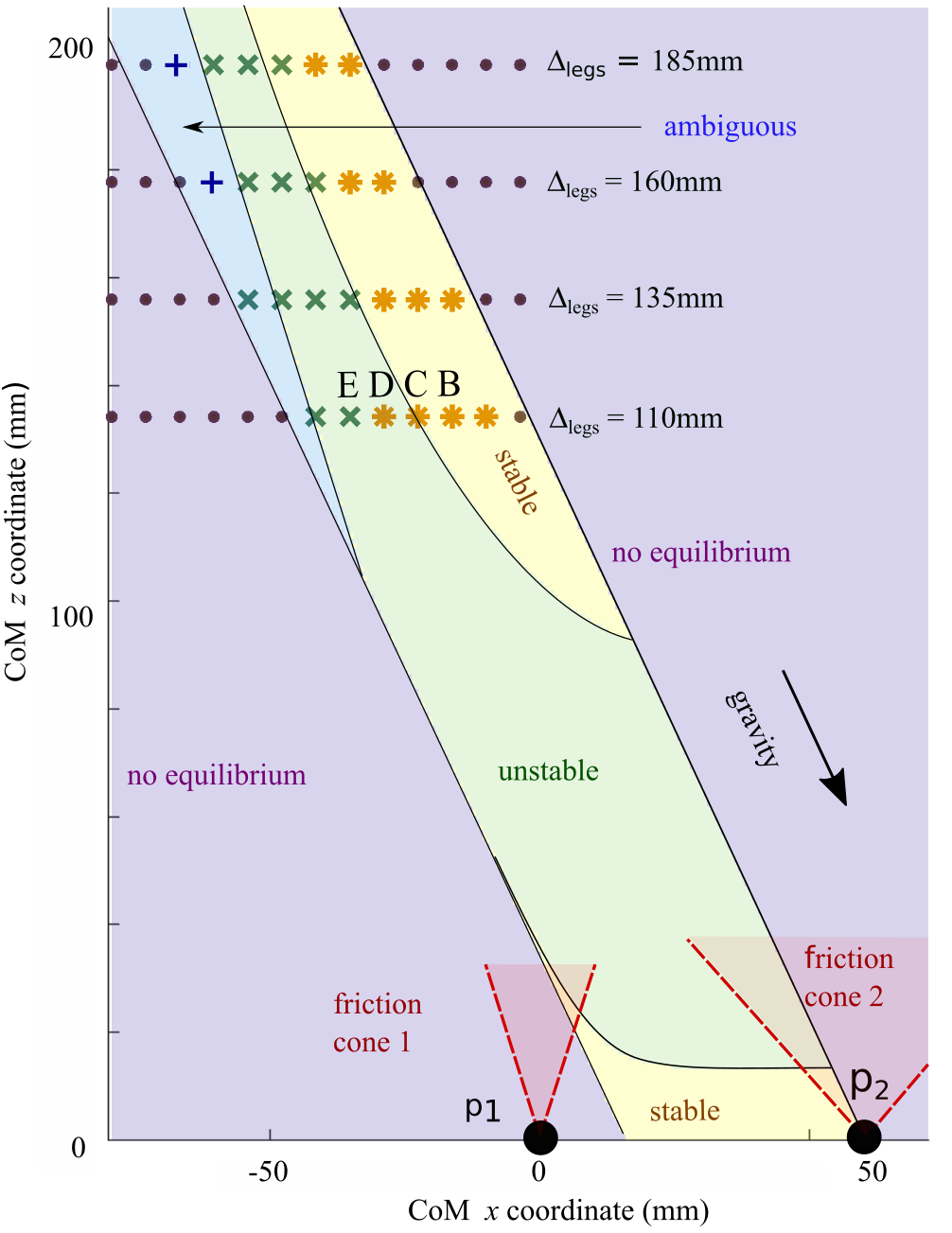}
\caption{Comparison of CoM stability regions, experiments vs. theory: Plot of the biped on slope with fixed contact points and variable CoM location. ($\{x,z\}$ axes are rotated to match tangential and normal directions of the plane, tilted arrow denotes direction of gravity). Color-shaded regions denote CoM locations for ambiguity-instability, reverse-chatter instability and stability, according to our theory. Discrete markers denote qualitative experimental characterization of the biped's response to perturbation: `$\bullet$' - no equlibrium. `$+$' - ambiguity-instability. `$\times$' - reverse-chatter instability. `*' - stability. Dashed lines denote friction cone boundaries emanating from the contact points $\vp_1,\vp_2$.} \label{fig_stability_plane}
\end{figure}

We conducted two sets of experiments. In the first set, we varied the legs' lengths $\Delta_{legs}$ and the location of one heavy cylinder while the spacing between the two legs was kept constant at distance of $60$~mm. By doing so, the biped's center-of-mass positions $(x_c,z_c)$ covered a discrete grid of $13 \times 5$ points, 
where the $\{x,z\}$ axes are \{aligned with, normal to\} the inclined slope. Each position was first tested for maintaining static equilibrium, and the feasible equilibrium postures   have been perturbed by slightly lifting and releasing one of the footpads. The biped's motion typically consisted of a sequence of alternating impacts at the two footpads, which have been recorded by a regular Smartphone camera (frame rate 30 fps) and classified as stable or unstable. Situations of highly sensitive equilibrium which tends to slide and tip over without further impacts were classified as ambiguous equilibria.

\begin{figure}
\centering
\includegraphics[width=0.44\textwidth]{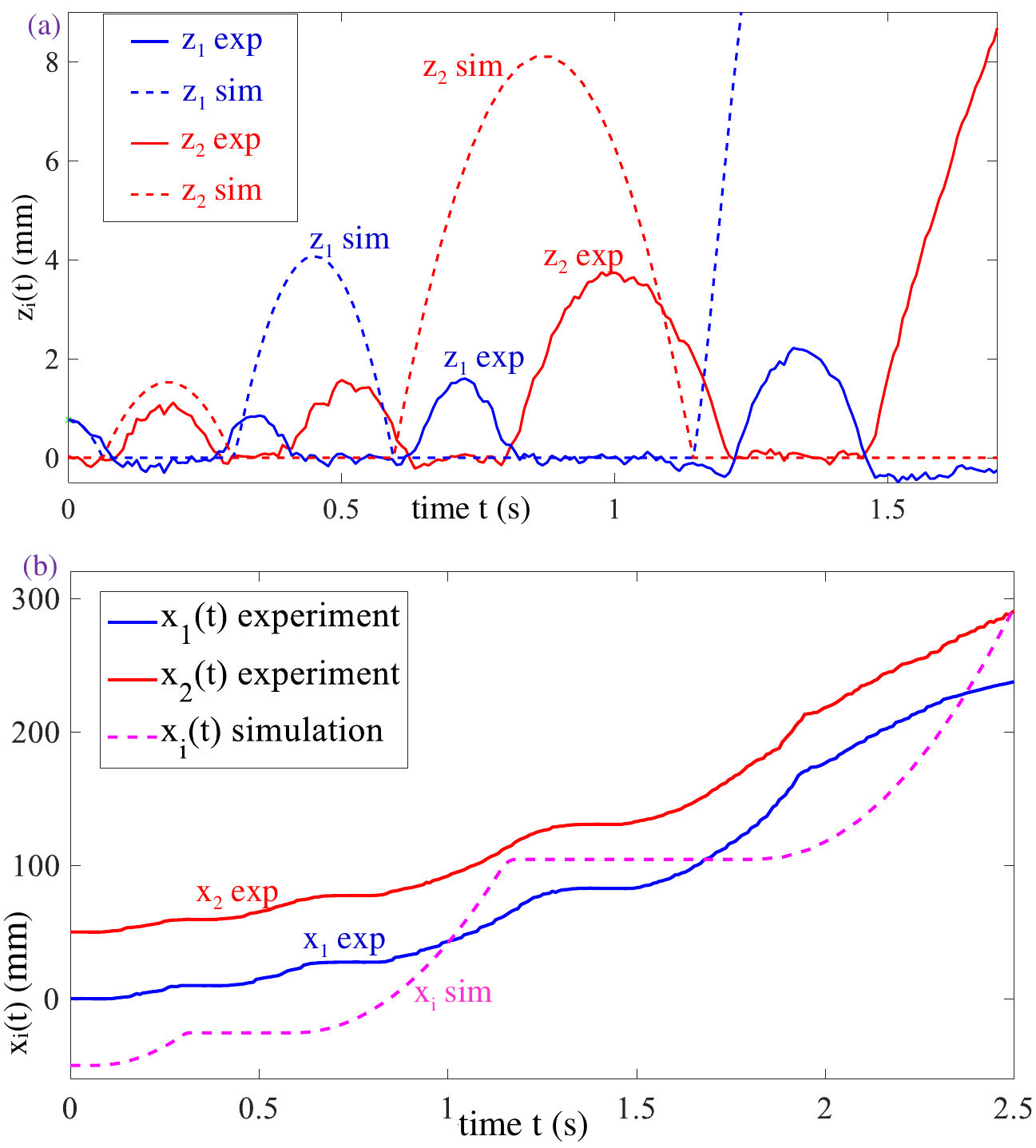}
\caption{Experiment E (unstable): comparison between experimental measurements (solid curves) and theoretical ZOD simulations (dashed). (a) Normal displacements $z_i(t)$. (b) Tangential displacements $x_i(t)$, with shifted initial values at $t=0$. 
 } \label{fig_expB}
\end{figure}

Figure \ref{fig_stability_plane} shows a grid of CoM positions of all experiments, marked according to their stability classifications. Markers `*' denote stable equilibrium while `$\times$' denote unstable ones and `$\bullet$' denote no feasible equilibrium. Two cases which were identified as ambiguous and unstable equilibrium are denoted by `$+$'. The discrete grid is overlaid on continuous color-shaded regions of \{infeasible, ambiguous, unstable, stable\} equilibrium, computed using our theoretical stability analysis as described in Section \ref{sec:newtheory}. (Note that shifting the weights also imposes changes in the biped's moment of inertia, which are also accounted in our ZOD analysis). One can observe a good qualitative agreement between theory and experiments. In particular, the experiments capture the main trend in stability changes: moving the CoM downhill along $+x$ direction (by shifting the heavy cylinder's mounting point along the beam) or up in $+z$ direction (by extending the legs' lengths) increases the stability. The exact values of stability transitions are highly sensitive to variations of the lower friction coefficient $\mu_1$ (see Fig. \ref{fig_expB} below for further illustration of this point).

\begin{figure}
\centering
\includegraphics[width=\columnwidth]{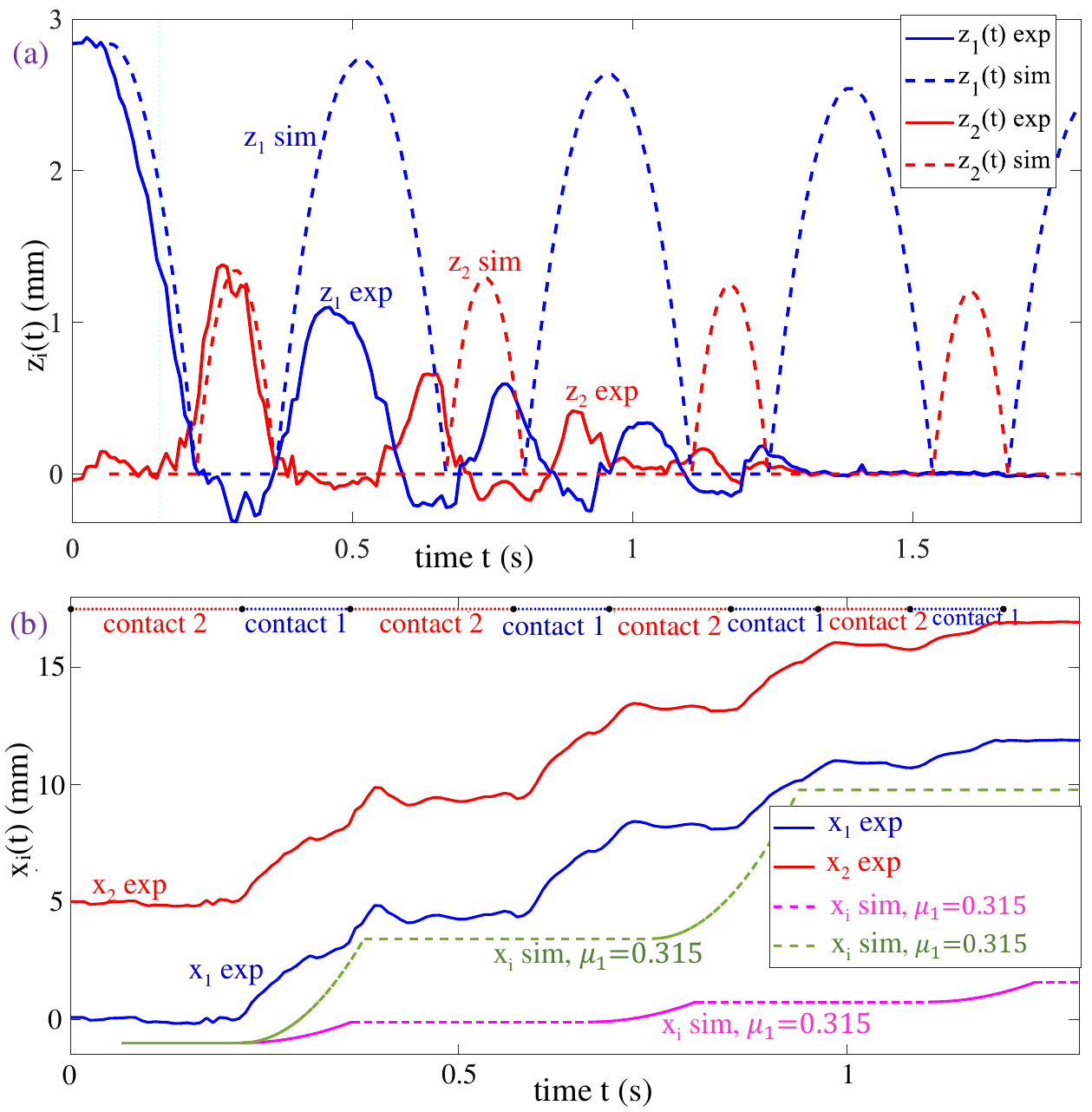}
\caption{Experiment C (stable): comparison between experimental measurements (solid curves) and theoretical ZOD simulations (dashed). (a) Normal displacements $z_i(t)$. (b) Tangential displacements $x_i(t)$, with shifted initial values at $t=0$.} \label{fig_expA}
\end{figure} 

In the second set of experiments, we used a fast camera (Samsung EX2F, frame rate 120 fps) for accurate tracking of the biped's motion due to perturbation from equilibrium, for selected cases of the biped's structure (mainly legs' lengths of $\Delta_{legs}=\{110$~mm$,135$~mm$\}$). Circular markers have been attached to the biped for motion tracking, see Figure \ref{fig_setup}b. Processing of the video movies of the experiments has been conducted using the commercial software TEMA \cite{TEMA}. Using rigid-body kinematics and known geometric relations, the footpads' positions have been calculated using the tracked markers' positions, and decomposed into their normal and tangential components $x_i(t),z_i(t)$, after calibration with respect to their values at two-contact equilibrium state. While the complete raw data appears in the supplementary material, we show here representative examples of data analysis for two specific experiments with legs length of $110$~mm and different CoM position $x_c$, which are denoted by C (stable point) and E (unstable point) in the grid of Fig. \ref{fig_stability_plane}. Time plots of the normal displacements $z_i(t)$ and tangential displacements $x_i(t)$ of the two footpads for configuration C are shown in Fig. \ref{fig_expA}, while results for configuration E are shown in Fig. \ref{fig_expB}. The dashed curves denote the results of ZOD simulations under the same initial conditions and geometric parameters of the biped. In the plots of $x_i(t)$ (Fig. \ref{fig_expA}b and \ref{fig_expB}b), the curves grow monotonically in time and have been separated vertically by adding different constant values in order to improve visibility. One can see excellent qualitative and moderate quantitative agreement between the theoretical ZOD-based simulation and the measured motion. Main observations 
are as follows. First, in Fig. \ref{fig_expA}a and \ref{fig_expB}a of $z_i(t)$, the simulations and experiments both capture the alternating order of footpads' contact-separation and the convergence (Fig. \ref{fig_expA}a) or divergence (Fig. \ref{fig_expB}a) of the bouncing sequences. Second, the restitution and growth ratio between consecutive impacts in the experimental measurements are clearly lower than those in the theoretical simulations. One clear reason for this is the small apparent  ``penetration'' $z_i<0$ which occurs in both footpads right after impacts ($z_i=0$ is calibrated at rest in two-contact equilibrium). This is caused by compression of the rubber hemispherical footpads, combined with elastic deflection of the slender legs. This effect contributes to increased energy dissipation beyond the losses due to rigid-body impacts predicted by the theoretical model. In addition, closer look on the curves of $z_i(t)$ of contact-free footpads reveals small superposed oscillations caused by elastic vibrations of the biped's slender structure (beam and legs), which are also not accounted by the rigid-body model.      
As for the tangential motion $x_i(t)$ in Figs. \ref{fig_expA}b and \ref{fig_expB}b, under ZOD model the tangential motion of both contact is identical (assuming negligible body rotations). The horizontal dotted line segments at the top of Fig. \ref{fig_expA}b denote alternating time intervals where each footpad is in contact. This illustrates the fact that downhill slippage occurs mainly when the upper footpad $\vp_1$ is in contact, since it has the lower friction. This is in agreement with the theoretical prediction (details of slippage timing for the simulated dashed curve are not shown). In Fig. \ref{fig_expA}b, we show simulation curves of $x_i(t)$ for two values of the smaller friction coefficient $\mu_1$, in order to illustrate the sensitivity of the slippage distance with respect to  variations in $\mu_1$. Finally, it can be seen that the total tangential displacement due to slippage in the unstable case of experiment E (Fig. \ref{fig_expB}b) is significantly larger than the stable case of experiment C (Fig. \ref{fig_expA}b). This agrees with the theoretical prediction that increased slippage during the cycle enables gaining more kinetic energy from drop in potential energy while slipping down the slope, which compensates for impact losses and frictional dissipation and leads to divergence of the bouncing sequence \cite{varkonyi2012lyapunov}.

\begin{figure}
\centering
\includegraphics[width=\columnwidth]{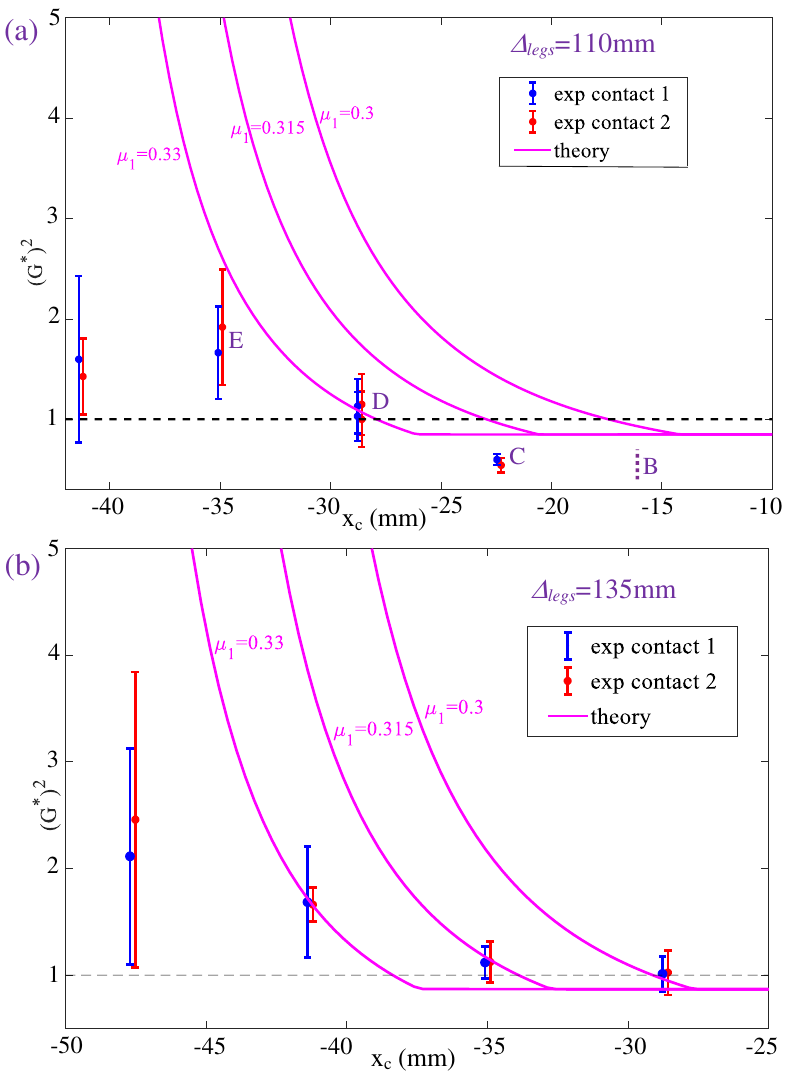}
\caption{The squared growth ratio $(G^*)^2$ as a function of horizontal CoM position $x_c$ for different leg lengths of (a) $\Delta_{legs}=110$~mm and (b) $\Delta_{legs}=135$~mm. Discrete dots with error bars are mean values calculated from measurements in experiments. The solid curves are from theoretical ZOD analysis. The dashed line of $(G^*)^2=1$ denotes the stability limit. The growth ratio for configuration B could not be determined due to the small stability region of the equilibrium} \label{fig_G}
\end{figure}
\begin{figure*}[h!]
\centering
\includegraphics[width=0.95\textwidth]{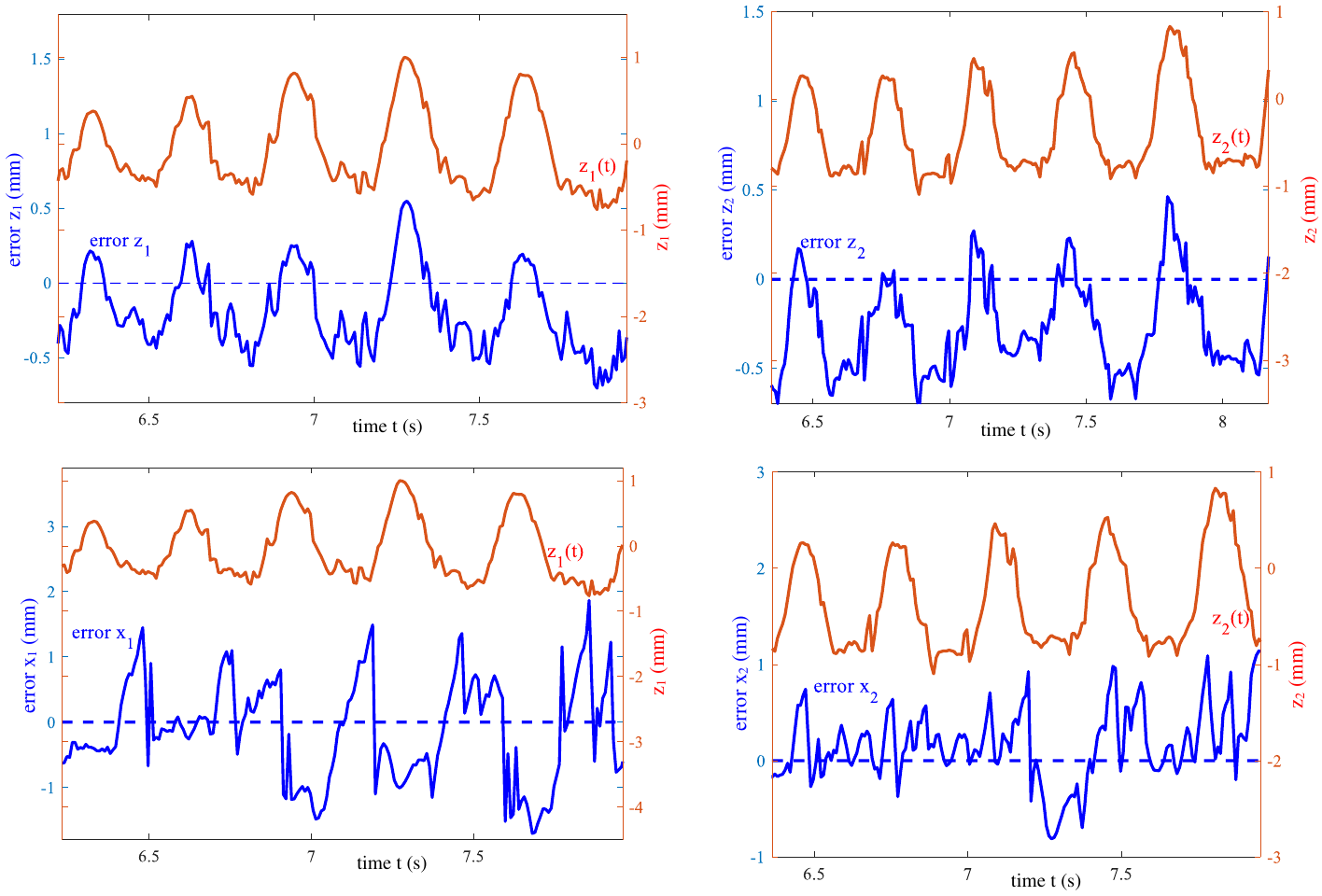}
\caption{Comparison of markers to footpads tracking for experiment D. Errors in $x_i(t)$ and $z_i(t)$ overlaid with $z_i(t)$.} \label{fig_footpads}
\end{figure*}

Next, we analyzed the bouncing sequences of measured normal displacements $z_i(t)$, and computed average ratio between consecutive peak points of $z_i(t)$ along the motion. These values are plotted in Fig. \ref{fig_G}a and \ref{fig_G}b as discrete markers plus standard deviation error bars for fixed legs' length $\Delta_{legs}$ and different positions of the movable weight along the beam. 
This measurement was done only for those configurations, where the stability region of the equilibrium was large enough to provide sufficiently long time series. For example, the growth ratio of configuration B in Fig. \ref{fig_G}a could not be measured. 
 
The measured results are compared with the theoretical computation of squared growth ratio $(G^*)^2$, plotted in Fig. \ref{fig_G}a-b as continuous curves. 
Three curves are plotted which are calculated under different values of the friction coefficient $\mu_1$, illustrating the high sensitivity of decay (growth) rate and the stability crossing point ($G^*=1$) with respect to $\mu_1$. The comparison again shows good qualitative agreement between the theory and experimental measurements, where discrepancies can be explained by the following observations. First, as stated above, the experimental biped is, in most cases, ``more stable'' than the theoretical system under the same physical values, showing lower decay  (or divergence) rate of the bouncing sequences. This is due to added energy dissipation contributed by compression of the footpads and elastic deformations, effects which are both not accounted by the theoretical rigid-body model. Second, the ZOD model ignores all nonlinear effects of the dynamics by evaluating all accelerations at the equilibrant position under zero velocities while neglecting all nonlinear effects of rotations and velocity-squared terms. Furthermore, the ZOD curve of $(G^*)^2$ accounts only for motion along the cyclic solution associated with the fixed point $\vfi^*$ of $R(\vfi)$, and thus ignores the transient response due to initial conditions which are away from $\vfi^*$.

Next, in attempt to quantify the deviations of the biped's structure from rigid shape, we compare direct tracking of the hemispherical footpads with the data obtained from tracking the markers and using kinematic rigid-body transformations for calculating footpads' positions. We used the software to locate the hemispheres as point clouds and calculate their geometric center. The calculations of footpads' positions using the two methods have been calibrated by setting a reference value of $\vp_i=0$ at two-contact equilibrium posture before applying the initial perturbation. We denote $f_{error}=f_{marker}-f_{footpad}$ where $f(\cdot)$ are contact displacements $x_i(t),z_i(t)$.  The results are shown in the plots in Fig. \ref{fig_footpads} for the case of experiment D (shown in the grid of Fig. \ref{fig_stability_plane}). This case was chosen since it is characterized by ``marginal stability'' where a long bouncing sequence has been observed with roughly constant magnitude without quick growth or decay, enabling more reliable data acquisition. Errors in displacements of the $i$-th contact are overlaid with the normal displacement $z_i(t)$ (using markers' data) in order to visually clarify the correlations
between errors and the footpad's contact state.  

Main observations from Fig. \ref{fig_footpads} are as follows. First, in the displacements $z_i(t)$, when the $i$-th footpad comes to contact there is a persistent negative error of ``contact penetration'' of up to $\sim 0.5$~mm. This seems to be a consequence of compression of the rubber footpad combined with elastic deflection of the biped's slender elements. Note that this compression also resulted in notable deformation of the rubber footpads from their nominal semi-circle image, which induced noises on direct footpads' tracking. In the tangential displacements $x_i(t)$, one can see larger deviations of up to $\sim 1$~mm, which may be caused by elastic bending of the legs. Another notable effect is the rapid jumps in the error of $x_i(t)$ from positive to negative, which occur right after contact release of the $i$-th footpad ($z_i(t)$ increasing). This can be explained by deflection of the slender leg when the body moves downhill and the footpad lags behind constrained by the contact, followed by quick release where the footpad ``catches up`` forward with some overshoot. Finally, one can clearly see superimposed oscillations in errors of both $x_i(t)$ and $z_i(t)$,  which may be caused by elastic vibrations of the biped's structure. Such oscillations are present both during contact and released phases of each footpad.

\section{Conclusion}
\label{sec:conclusions}

In this work, we have presented for the first time an experimental demonstration of our previous theory of Lyapunov stability for a planar rigid body with two frictional contacts \cite{varkonyi2017lyapunov}. Our chosen example of biped on a slope with two rubber footpads having unequal friction, results in crucial dependence of stability on the CoM location, which can be manipulated by varying the internal structure of the body. We presented extension of our stability theory in \cite{varkonyi2017lyapunov} in order to cover cases of weakly-persistent equilibria, and derived a simple stability criterion based on fixed point $\varphi^*$ of the reduced \Poincare map $R(\varphi)$ and their growth values $G(\varphi^*)$. The theoretical predictions have been corroborated by experiments, clearly showing changes of stability of the alternating impact sequences, from stable Zeno convergence to unstable divergence via reverse chatter. Using fast video recording and motion tracking enabled detailed analysis of the motion and testing validity of the rigid-body model. We found remarkable agreement with numerical simulations of the zero-order dynamics under rigid-body model, where some deviations are caused by compression of the rubber footpads and elastic vibrations of slender parts of the structure. These effects add energy dissipation which tends to slightly increase stability relative to prediction of the rigid-body model. Nevertheless, good agreement between theory and experimental results is achieved, and the dependence of stability on CoM position is clearly captured.

We now briefly discuss limitations of our current work and suggest possible directions for future extension of the research. First, our stability theory is currently limited to a rigid body. Extension to stability analysis of frictional equilibria of robots composed of multi-body links will enable incorporation of feedback control laws into actuated joints in order to induce Lyapunov stability. While computational aspects of this direction are partially covered in \cite{posa.stability.tac2016}, we wish to find low-dimensional models which are amenable to semi-analytic investigation. This may help in gaining insights 
into basic mechanisms of stabilization. We also wish to extend the theory to incorporate more than two contacts and three-dimensional motion. Some initial steps in this direction have been taken in \cite{baranyai2018zeno}. \red{Nevertheless, in the case of many contact points the complexity of the hybrid dynamics approach generated by the exponentially growing number of contact modes calls for a different approach. These efforts may lead towards} experimental realization of feedback stabilization of frictional equilibria in multi-contact robotic systems based on hybrid-dynamics theoretical analysis, which is a major open challenge for longer term.

Finally, the result of the stability analysis is often highly sensitive to model parameters such as the friction coefficent, which are not accurately known in practical situations. In such cases, the stability analysis must be combined with sensitivity analysis in order to develop robust and reliable stability tests. 
\red{The effect of unmodelled aspects of contact dynamics such as elastic deformations, partially elastic impact and rolling friction \cite{acary2021coulomb} may also require further clarification.}

%

%


\appendices

\section{The equations of motion }
{
First we introduce the notation
\begin{align*}
\eta_i&=l_i\cos\phi_i-h\sin\phi_i\\
\xi_i&=h\cos\phi_i+l_i\sin\phi_i
\end{align*}
\begin{align*}
\zeta_i&=\xi_i\sin\theta+\eta_i\cos\theta\\
\psi_i&=\xi_i\cos\theta-\eta_i\sin\theta
\end{align*}
The elements of $\mathbf{q}$ can be expressed as 
\begin{align}
z_1&=-x\sin\phi_1+z\cos\phi_1+\xi_1-\psi_1  \\
z_2&=-x\sin\phi_2+z\cos\phi_2+\xi_2-\psi_2       \\
x_2&=x\cos\phi_2+z\sin\phi_2-\eta_2+\zeta_2  
\end{align}
Taking the second derivative of these three equations with respect to time, and substitution of \eqref{eqcontinuousmotion}  yields an expression for $\ddot{\mathbf{q}}$ expressed in terms of $\theta$, $\dot\theta$, and the unknown contact forces:
\red{
\begin{align}\label{eq:continuousmotion-new}
m\ddot{\mathbf{q}}&=
f_{ex}\begin{bmatrix}
-\cos(\alpha-\phi_1)\\-\cos(\alpha-\phi_2)\\\sin(\alpha-\phi_2)
\end{bmatrix}
+
\rho^{-2}\tau_{ex}\begin{bmatrix}
\zeta_1\\
\zeta_2\\
\psi_2
\end{bmatrix}
+
m\dot\theta^2\begin{bmatrix}
\psi_1\\
\psi_2\\
-\zeta_2
\end{bmatrix}\nonumber\\
&...+\sum_{i=1}^2 
f_i^{z}
\begin{bmatrix}
\cos(\phi_1-\phi_i)+\rho^{-2}\zeta_i\zeta_1\\
\cos(\phi_2-\phi_i)+\rho^{-2}\zeta_i\zeta_2\\
\sin(\phi_2-\phi_i)+\rho^{-2}\zeta_i\psi_2\\
\end{bmatrix}
\nonumber\\
&...+\sum_{i=1}^2 
f_i^{z}
\begin{bmatrix}
\sin(\phi_i-\phi_1)+\rho^{-2}\psi_i\zeta_1\\
\sin(\phi_i-\phi_2)+\rho^{-2}\psi_i\zeta_2\\
\cos(\phi_i-\phi_2)+\rho^{-2}\psi_i\psi_2\\
\end{bmatrix}
\end{align}
}
The ZOD approximation is obtained by substituting $\theta=\dot
\theta=0$:
\red{
\begin{align}
m\ddot{\mathbf{q}}&=
f_{ex}\begin{bmatrix}
-\cos(\alpha-\phi_1)\\-\cos(\alpha-\phi_2)\\\sin(\alpha-\phi_2)
\end{bmatrix}
+
\rho^{-2}\tau_{ex}\begin{bmatrix}
\eta_1\\
\eta_2\\
\xi_2\\
\end{bmatrix}\nonumber\\
&...+\sum_{i=1}^2
f_i^{z}
\begin{bmatrix}
\cos(\phi_1-\phi_i)+\rho^{-2}\eta_i\eta_1\\
\cos(\phi_2-\phi_i)+\rho^{-2}\eta_i\eta_2\\
\sin(\phi_2-\phi_i)+\rho^{-2}\eta_i\xi_2\\
\end{bmatrix}
\nonumber\\
&...
+\sum_{i=1}^2
f_i^{x}
\begin{bmatrix}
\sin(\phi_i-\phi_1)+\rho^{-2}\xi_i\eta_1\\
\sin(\phi_i-\phi_2)+\rho^{-2}\xi_i\eta_2\\
\cos(\phi_i-\phi_2)+\rho^{-2}\xi_i\xi_2\\
\end{bmatrix}
\label{eq:ZODdetailed}
\end{align}
}
Note that this equation is independent of all state variables. Together with the equality constraints of individual contact modes, \eqref{eq:ZODdetailed} can be used to determine a constant acceleration value $\ddot{\mathbf{q}}$ in each contact mode.}
\section{Steps of the proof of Theorem \ref{thm.stability_general} requiring persistence}
 The proof of Theorem \ref{thm.stability_general} was given by \cite{varkonyi2017lyapunov} and it used a sequence of lemmas and definitions. Below, we make a list of those ones, which become  invalid or useless when the requirement of \emph{persistent equilibrium} is relaxed.. Then in the next section we will present modified statements along with the outlines of updates proofs, which are necessary to extend Theorem \ref{thm.stability_general} into Theorem \ref{thm.stability_general2}. 

First, two pseudo-metrics closely related to \eqref{eq.Delta} are defined as follows:
\begin{align}
D(\vq,\vqd)&=\max\left(\sqrt{z_{1}},\sqrt{z_{2}},|\dot{z}_{1}|,|\dot{z}_{2}|,|\dot{x}_{2}|\right),
\\
d(\vq,\vqd)&=\max\left(\sqrt{z_{1}},\sqrt{z_{2}},|\dot{z}_{1}|,|\dot{z}_{2}|\right).
\end{align}

Then, the first statement establishes an important property of the dynamics: that it must often undergo contact mode transitions and/or impacts:
\begin{lemma}[\cite{varkonyi2017lyapunov}]
\label{lem:finite-impacts}
For a given persistent equilibrium configuration of a planar rigid body on two unilateral frictional contacts, there exist finite positive scalars $k_{1}$ and $c_{1}$ such that any possible solution trajectory under the {\rm ZOD} assumption must satisfy the following bounds:\\
(i) if the system undergoes an impact at time $t_{1}$ then
\begin{equation}
D(t_{1}^{+})<k_{1}\cdot D(t_{1}^{-}) \mbox{ and }
d(t_{1}^{+})<k_{1}\cdot d(t_{1}^{-})
\label{eq:impact-Ddlimit} 
\end{equation}
(ii) if the systems undergoes no impact or contact mode transition
between times $t_{1}$and $t_{2}$, then
\begin{equation}
D(t)<k_{1}\cdot D(t_{1}) \mbox{ and } d(t)<k_{1}\cdot d(t_{1})
\, for\, all\, t_{1}<t<t_{2}. 
\label{eq:lemma-Ddlimit}
\end{equation}
(iii) in addition, if the systems is not in {\rm SS} mode at $t_{1}$, then
\begin{equation}
t_{2}-t_{1}\leq c_{1}\cdot D(t_{1}),\label{eq:lemma-tlimit for D}
\end{equation}
and if it is not in {\rm SS, PP, NN, PN}, or {\rm NP} mode then
also
\begin{equation}
t_{2}-t_{1}\leq c_{1}\cdot d(t_{1}).\label{eq:lemma-tlimit for d}
\end{equation}

\end{lemma}
The next statement characterized the behavior of $R$ and $G$ near their endpoints:
\begin{lemma}[Lemma 3 of \cite{varkonyi2017lyapunov}]
\label{lem:properties_RG} Assume that the maps $R(\vfi)$, $G(\vfi)$ associated with a persistent equilibrium configuration are defined at an endpoint $\vfi \to \pm \pi/2$. Then,
there exists a finite-sized sub-interval $(-\pi/2,\beta_0]$ or $[\beta_0,\pi/2)$ for which the growth map $G(\vfi)$ attains a constant value of
\beq{G.pi2}
G(\vfi)=G^\pm
\eeq
furthermore $R(\vfi)$ satisfies the following relations:
\beq{R.pi2} \lim_{\vfi \to \pm \tfrac{\pi}{2}}R(\vfi) = \pm \tfrac{\pi}{2} \eeq
\beq{dR.pi2} \lim_{\vfi \to \pm\tfrac{\pi}{2}}R'(\vfi) =  G^{\pm}, \mbox{  where $R'(\vfi)=dR/d \vfi$.}\eeq
\end{lemma}
Next, we present a sequence of definitions culminating in the definition of stable partitions  lying at the core of Theorem \ref{thm.stability_general}:
\begin{defn}
The {\em \bf partitioning} induced by an arbitrary finite collection of scalars $C:\lbrace -\pi/2 < \beta_1 < \beta_2 <...< \beta_r < \pi/2 \rbrace$ is the set of intervals $I_1=(-\pi/2,\beta_1],I_2=[\beta_1,\beta_2],\ldots, I_{r+1}=[\beta_{r},\pi/2)$ 
\end{defn}
\begin{defn}
The {\em \bf interval graph} associated with a partitioning and the map $R$ is a graph with $r$ vertices corresponding to the $r$ intervals. It has a directed edge $I_j\to I_k$ if $R(\varphi) \in I_k$ for some $\varphi \in I_j$. The graph does not have self-edges $(I_k \to I_k)$. 
\end{defn}
\begin{defn}
$I_1$ and $I_r$ are called {\em \bf extremal intervals} 
\end{defn}
\begin{defn}
An interval $I_k$ is called {\em \bf safe} if $G(\vfi)<1$ for all $\vfi \in I_k$
\end{defn}
\begin{defn}
An interval $I_k$ is called {\em \bf transient} if $R(\vfi)-\vfi$ is either strictly positive or strictly negative for all $\vfi \in I_k$, and additionally it does not belong to any closed loop of the interval graph 
\end{defn}
\begin{defn} \label{def:safepartition}
A partitioning associated with a persistent equilibrium is called a {\em \bf stable partition} if each interval is safe and/or transient, furthermore $G^\pm\neq 1$.
\end{defn}
Finally, recall that upper indices like $\vfi^{(k)}$, $t^{(k)}$ denote values of variables at the time of crossing the \Poincare section for the $k^{th}$ time. We use this notation to review  a lemma from \cite{varkonyi2017lyapunov} establishing an important consequence of the existence of stable partitions:
\begin{lemma} [Lemma 5 of \cite{varkonyi2017lyapunov}]
\label{lem:special intervals} For a two-contact persistent equilibrium configuration, if a stable partition exists and
$G^{+}>1$, ($G^{-}>1$),  then the extremal interval(s) in the partition can be chosen so that
there exist constants $c_{ex}$ and $k_{ex}$ such that any solution that 
    satisfies 
$\varphi^{(m)},\varphi^{(m+1)},...,\varphi^{(m+K)}\in I_{r}$
($I_{1}$),  is bounded by 
\begin{equation}
D^{(i)}\leq k_{ex}\cdot D^{(m)}\label{eq:Dlimit-extremal}
\end{equation}
for all $i=m,m+1,\ldots,m+K+1$, and 
\begin{equation} t^{(m+K+1)}-t^{(m)} 
\leq c_{ex}\cdot D^{(m)}\label{eq:time limit-extremal}
\end{equation}
\end{lemma}

\section{Modified statements and definitions}
Here we list all the essential additions that extend the main results of \cite{varkonyi2017lyapunov} to non-persistent but  weakly persistent two-contact equilibrium configurations.
\subsection{Extension of Lemma \ref{lem:finite-impacts}}

The statement of the lemma remains true in the absence of persistence, however the proof given by \cite{varkonyi2017lyapunov} requires some modification at one point. In step 4 of the original proof, we exploited the inequality 
\begin{equation}
{f}_i^z<0\textrm{ or }\ddot{z}_{j}<0 \label{eq:zdotdotsingleslip}
\end{equation} for any contact mode $\mathcal{M}$ within the set $\lbrace PF, NF, FP,$ $FN \rbrace$ and for $i,j$ being the indices of the slipping and of the free-flying point respectively. This was a consequence of persistence. Here we only assume unambiguity. For any unambiguous equilibrium, the mode $\mathcal{M}$ is not consistent in a state of rest, thus either \eqref{eq:zdotdotsingleslip} holds or 
\begin{equation}
-\sign (\dot{x}_{j})\ddot{x}_{j}<0 \label{eq:xdotdotsingleslip}
\end{equation}
Eq. \eqref{eq:xdotdotsingleslip} can be used in the proof in the same way as \eqref{eq:zdotdotsingleslip}.

\subsection{Extension of Lemma \ref{lem:properties_RG}}

In the case of non-persistent systems, the statement in Lemma \ref{lem:properties_RG} does not hold. Instead, the behavior of the $R$ and $G$ functions can be characterized by the following statement

\begingroup
\renewcommand\thelemma{\ref{lem:properties_RG}a}
\begin{lemma}
\label{lem:properties_RG_2} Assume that the maps $R(\vfi)$, $G(\vfi)$ are defined at an endpoint $\vfi \to \pm \pi/2$, and the equilibrium is unambiguous. Then
both endpoints $\varphi=\pi/2$ and $\varphi=-\pi/2$ of the $R$ and $G$ functions satisfy one of the following two sets of properties:\\
Set 1: the properties given by Lemma \ref{lem:properties_RG} above\\
Set 2:
%
	the reduced \Poincare  map $R(\vfi)$ converges to a constant value 
\beq{R.pi2:painleve}
\lim_{\vfi \to \pm \tfrac{\pi}{2}}R(\vfi)=R^\pm
\eeq
with $|R^{\pm}|<\pi/2$. Furthermore the growth map $G(\vfi)$ diverges to infinity as given by the relation

\beq{G.pi2.painleve} \lim_{\vfi \to \pm \tfrac{\pi}{2}}G(\vfi)(\pm\pi/2-\vfi) = \tan(R^\pm) \eeq
\end{lemma}
\endgroup
\begin{proof}
The statement specifies two sets of properties, one of which is claimed to hold for any system and limit point ($\varphi\rightarrow \pi/2$ or $\varphi\rightarrow -\pi/2$). In \cite{varkonyi2017lyapunov}, we already showed that Set 1 is true for any persistent equilibrium and for any of the two limit points. The proof was based on the observation that a crossing of the \Poincare section with $\varphi$ sufficiently near to that limit is followed by a full cycle of motion in which the only contact interactions are slipping in one given direction.

For non-persistent systems, the proof developed for persistent equilibria remains valid for $\vfi\rightarrow+\pi/2$ as long as \eqref{eq:zdotdotsingleslip} is true for those two contact modes, which may follow immediately after a Zeno point with $\vfi\rightarrow+\pi/2$ (specifically: $FP$ and either $PF$ of $NF$ depending on $\sign(\sin\phi_1\sin\phi_2)$). Similarly, the proof works in the case of $\vfi\rightarrow-\pi/2$ as long as \eqref{eq:zdotdotsingleslip} is true for the other two contact modes.

In what follows, we show that set 2 is satisfied in the opposite case. We discuss one out of four possible cases in detail but the other three are treated in an analogous way. Specifically, assume now that 
\begin{equation}
\ddot{z}_{1}^{(FP)}>0\label{eq:z1dotdotFP}
\end{equation} 
and consider a crossing of the \Poincare section with $\varphi\approx+\pi/2$. First note that this state corresponds to $z_1=z_2=\dot z_1=0$, $\lim_{\vfi\rightarrow\pi/2}\dot z_2=0$, $\dot x_2>0$, i.e. it is uniquely determined up to to a scalar scaling factor (the magnitude of $\dot x_2$). What follows is an immediate slipping impact (with vanishing velocity jump in the limit of $\vfi\rightarrow\pi/2$). If that impact is a double-contact impact, then $R$ is undefined, which is outside the scope of the lemma. Otherwise, $FP$ mode follows after the impact. In FP mode, point 1 accelerates away from the support surface by \eqref{eq:z1dotdotFP}. Hence, the only way for the FP mode to terminate is that tangential motion at point 2 stops due to \eqref{eq:xdotdotsingleslip} giving rise to stick or slip reversal. In the state when point 2 stops, the values of the state variables $z_1$, $z_2$, $\dot z_1$, $\dot z_2$, $\dot x_2$ are again determined up to a scalar scaling factor, which follows from the analogous property of the initial state. Furthermore, none of the five values vanishes in the limit $\vfi\rightarrow\pi/2$. From this observation, it can be shown that the value of $\vfi$ upon the next crossing of the \Poincare section is determined uniquely and $\vfi\neq\pm\pi/2$. Hence \eqref{eq.R.pi2:painleve} is satisfied. Similar arguments apply if \eqref{eq:zdotdotsingleslip} fails for any other contact mode. This completes the proof of \eqref{eq.R.pi2:painleve}.   

The proof of \eqref{eq.G.pi2.painleve} is then obtained as the result of straightforward calculation, following the steps of \cite{varkonyi2017lyapunov}. The interested reader should consult the section entitled "Proof of equation (31) in Lemma 3" in the Appendix of that paper. The detailed calculation is omitted here.
\end{proof}

\subsection{Improved definition of stable partitions}
The original definition of a stable partition used the values $G^\pm$. According to Lemma \ref{lem:properties_RG_2}, $G^\pm$ exists for some non-persistent equilibria (see set 1 of conditions in the lemma), but it does not exist for others (see set 2, which includes $\lim_{\vfi\rightarrow\pm\pi/2}G(\vfi)=\infty$). Accordingly, the definition requires a minor addition:
\begingroup
\renewcommand\thedefinition{\ref{def:safepartition}a}
\begin{definition} \label{def:safepartition2}
A partitioning associated with a weakly persistent equilibrium is called a {\em \bf stable partition} if each interval is safe and/or transient, and additionally if any of $G^\pm$ exists then it satisfies $G^\pm\neq 1$.
\end{definition}
\endgroup

\subsection{Extension of Lemma \ref{lem:special intervals}}
Finally, we need to extend  the proof of Lemma \ref{lem:special intervals} in order to cover all possible cases:
\begingroup
\renewcommand\thelemma{\ref{lem:special intervals}a}
\begin{lemma}
\label{lem:special intervals2} For a two-contact weakly persistent equilibrium configuration, if a stable partition exists  then the extremal interval(s) in the partition can be chosen so that
there exist constants $c_{ex}$ and $k_{ex}$ such that any solution that 
    satisfies 
$\varphi^{(m)},\varphi^{(m+1)},...,\varphi^{(m+K)}\in I_{r}$
($I_{1}$),  is bounded by 
\begin{equation}
D^{(i)}\leq k_{ex}\cdot D^{(m)}\label{eq:Dlimit-extremal2}
\end{equation}
for all $i=m,m+1,\ldots,m+K+1$, and 
\begin{equation} t^{(m+K+1)}-t^{(m)} 
\leq c_{ex}\cdot D^{(m)}\label{eq:time limit-extremal2}
\end{equation}
\end{lemma}
\endgroup

\begin{proof}
The case when both endpoints of the $R$ map follows Set 1 of conditions given in Lemma \ref{lem:properties_RG_2} has been discussed in \cite{varkonyi2017lyapunov}. Here we only need to dicuss what happens when one or both endpoints of $R$ follow Set 2.

As a preliminary step, for those endpoints which follow Set 2, we examine whether the corresponding extremal interval overlaps with the closed interval $(\min_\varphi R(\varphi),\max_\varphi R(\varphi))$. If they overlap, we cut the extremal interval into two smaller, nonempty intervals, such that the reduced extremal interval does not overlap with $(\min_\varphi R(\varphi),\max_\varphi R(\varphi))$.  This is possible because $(\min_\varphi R(\varphi),\max_\varphi R(\varphi))$ is bounded away from $\pm\pi/2$ according to \eqref{eq.R.pi2:painleve} in Lemma \ref{lem:properties_RG_2} and the extreme value theorem. We also revise the 'safe'/'transient' labels of the intervals affected by the cut. Such a cut preserves the stable property of the partitioning.

As a result of the cut, the $R$ map does not map any $\vfi$ into any of the extremal intervals where $R$ follows Set 2 of properties. Consequently, we may only have $m=1$, $K=0$ in Lemma \ref{lem:special intervals2}. Then, the system may only undergo a bounded number of contact mode transitions and impacts before these extremal intervals are left. Then Lemma \ref{lem:finite-impacts} implies \eqref{eq:Dlimit-extremal2}, and \eqref{eq:time limit-extremal2} in a straightforward way.
\end{proof}

\subsection{Proof of Theorem \ref{thm.final}}
The proof is based on constructing a stable partition of $I$ under the conditions outlined in the statement. 
Let  $\vfi_i^*$ $i=1,2,...,k$ denote the set of fixed points of $R$. We consider  the partition induced by the values $\vfi_1^*-\varepsilon,\vfi_1^*+\varepsilon,\vfi_2^*-\varepsilon,\vfi_2^*+\varepsilon,...,\vfi_k^*+\varepsilon$ where $\varepsilon$ is a sufficiently small positive scalar. We have $G(\vfi_i^*)<1$. By continuity of $G$, it is possible to choose $\varepsilon$ such that $G(\vfi)<1$ over every interval containing a fixed point, i.e. $(\vfi_i^*-\varepsilon,\vfi_i^*+\varepsilon)$. Hence all of these intervals are safe. All remaining intervals are transient because the transition graph associated with a non-decreasing map may not contain any cycles. Hence the partition that we have constructed is a stable one and Theorem \ref{thm.stability_general2} implies Theorem \ref{thm.final}. \qed

\section*{Acknowledgment}
The authors wish to deeply thank Ms. Ortal Halevi-Tabachnik and Dr. Yossi Elimelech for their devoted work on image processing, motion tracking and data analysis from video recordings of the experimental results, Benj\'amin Savanya for creating video recording, Ott\'o Sebesty\'en for building the test rig, as well as Dr. David Gontier, Prof. Joel W. Burdick, and 3 students of California Institute of Technology for building and testing a preliminary prototype of the experimental setup. PLV acknowledges financial support from the National Research, Innovation and Development Office of Hungary under grant $K124002$. YO acknowledges support from the Israel Science Foundation, under grant no. $1005/19$.

\ifCLASSOPTIONcaptionsoff
  \newpage
\fi



\bibliography{references,elonbib,lifebib,books,yizhar_bib}
\bibliographystyle{ieeetr}

 
\begin{IEEEbiography}[{\includegraphics[width=1in,height=1.25in,clip,keepaspectratio]{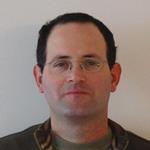}}]{Yizhar Or}
received the Ph.D. degree in mechanical engineering from Technion-Israel Institute of Technology, Haifa, Israel, in 2007. He is currently an Associate Professor of Mechanical Engineering with Technion-Israel Institute of Technology. His research interests include dynamics and control of robotic locomotion, analysis of nonholonomic underactuated robots, and hybrid dynamics of robotic systems with intermittent frictional contacts.
\end{IEEEbiography}
\newpage
\begin{IEEEbiography}[{\includegraphics[width=1in,height=1.25in,clip,keepaspectratio]{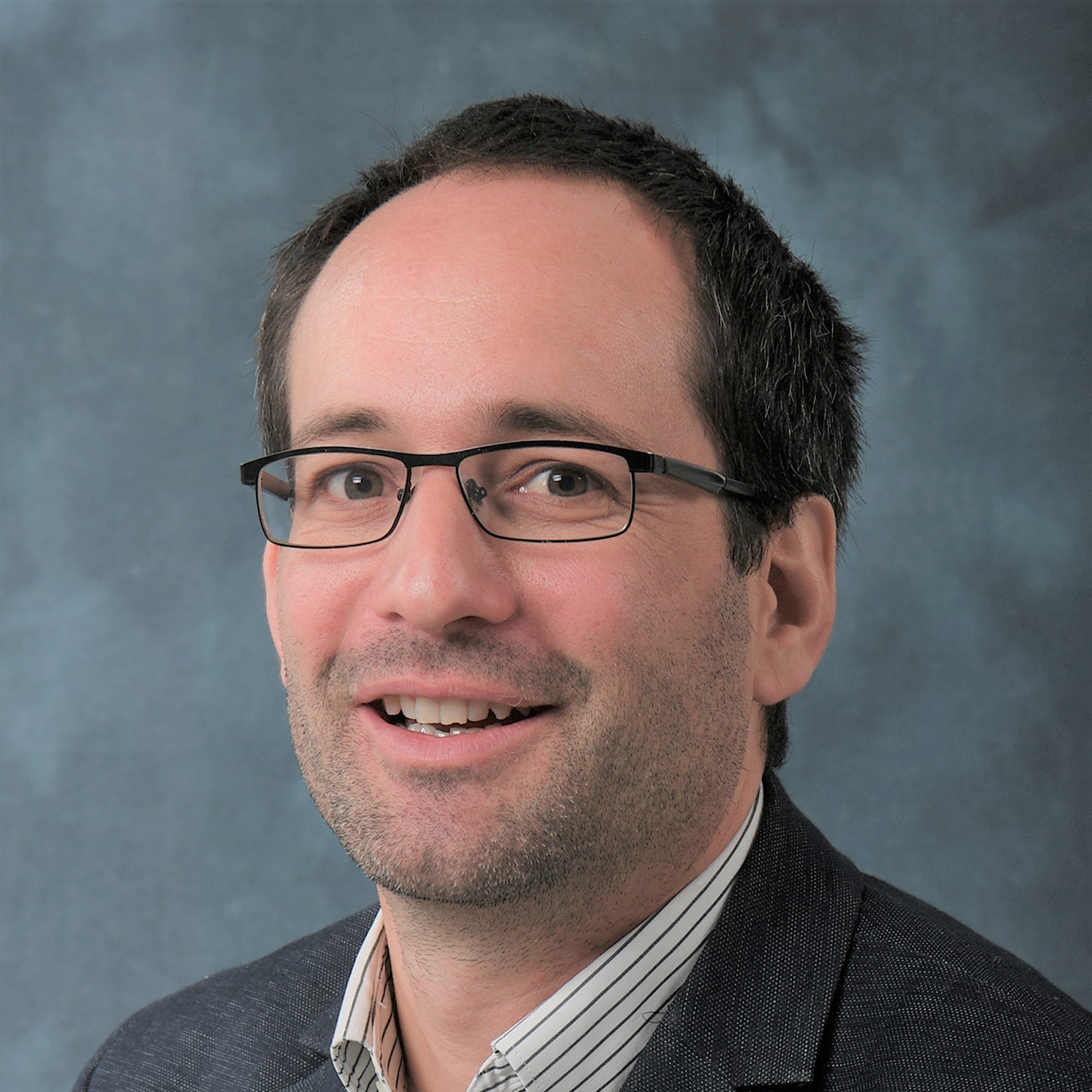}}]{P\'eter L. V\'arkonyi}
received the Ph.D. degree in architectural engineering from Budapest University of Technology and Economics, Hungary, in 2006. He is currently a Professor of Applied Mechanics with Budapest University of Technology and Economics. His research interests include nonsmooth, nonlinear and hybrid dynamics, and its application in robotics, mechanical and structural engineering, as well as biomechanics.
\end{IEEEbiography}




\vfill


\end{document}